%% file: main.tex
\documentclass[10pt]{article} 
\usepackage[accepted]{tmlr}

\input{math_commands.tex}

\usepackage{microtype}
\usepackage{graphicx}
\usepackage{booktabs} 

\usepackage{hyperref}
\usepackage{paralist}

\usepackage{hyperref}
\usepackage{url}

\usepackage{amsmath}
\usepackage{amssymb}
\usepackage{mathtools}
\usepackage{amsthm}

\input{math_commands.tex}
\hypersetup{hidelinks}
\usepackage{hyperref}
\usepackage{url}

\usepackage{booktabs}
\usepackage{tabularx}
\usepackage{siunitx}
\usepackage{makecell}
\usepackage{threeparttable}
\usepackage{multirow} 
\usepackage{array,xcolor}
\usepackage{amssymb}
\usepackage{graphicx}
\usepackage{subcaption}
\usepackage{caption}
\usepackage{wrapfig}   

\usepackage[table]{xcolor}
\usepackage[capitalize,noabbrev]{cleveref}

\usepackage{xcolor}

\newcommand{\revision}[1]{#1}

\theoremstyle{plain}

\theoremstyle{definition}

\theoremstyle{remark}

\usepackage[textsize=tiny]{todonotes}
\newcommand{\methodname}{\textsc{RMB}}
\title{\methodname: Reward Model Boosting Mitigates Reward Hacking}

\author{\name Jiabin Fan$^{1}$ \email jiabin@ualberta.ca\\
\name Dezhi Ye$^{2}$ \email dezhiye@tencent.com\\
\name Yongchang Hao$^{1}$ \email yongchang1@ualberta.ca\\
\name Lili Mou$^{1,3}$ \email doublepower.mou@gmail.com\\[0.5em]
\addr $^{1}$ Dept.~Computing Science \& Alberta Machine Intelligence Institute (Amii), University of Alberta\\
\addr $^{2}$ Tencent\\
\addr $^{3}$ Canada CIFAR AI Chair\\[0.4em]
}

\def\month{09}  
\def\year{2026} 
\def\openreview{\url{https://openreview.net/forum?id=bIaCQSyb4g}} 

\begin{document}

\maketitle

\begin{abstract}
Reinforcement Learning from Human Feedback (RLHF) is a powerful technique for aligning large language models (LLMs) with human preference. However, it often suffers from the reward hacking issue, where policy optimization improves the proxy reward model while actually degrading performance
with respect to the true human preference, due to the imperfection of the proxy. To address this, we propose Reward Model Boosting (\methodname), a novel approach that enhances the robustness and reliability of the reward signal for RLHF. \methodname{} first trains a set of reward models with a diversity-promoting regularizer. This encourages each model to learn complementary aspects of the reward landscape. Then, \methodname{} learns a lightweight aggregator in the principle of boosting to aggregate the outputs of the diverse reward models into a more accurate and robust reward signal. Our extensive experiments demonstrate that \methodname{} significantly improves reward accuracy on both in-distribution and out-of-distribution datasets, substantially mitigating the reward hacking issue and ultimately improving RLHF performance.\footnote{Our code is released at \url{https://github.com/MANGA-UOFA/RMB}}
\end{abstract}

\section{Introduction}

Large language models (LLMs) can be guided to follow human preferences through reinforcement learning from human feedback (RLHF), which utilizes a reward model trained on labeled human preference data to approximate human judgments of output quality \citep{bai2022training, ouyang2022training, achiam2023gpt, touvron2023llama}. By optimizing the LLM (i.e., the policy) towards this reward model, we can significantly improve the performance of the outputs. This approach has been widely adopted in recent LLM development and has demonstrated strong empirical success in aligning model behavior with human preferences \citep{ouyang2022training, team2024gemma, qwen2025qwen25technicalreport}.

However, a learned reward model for RLHF often suffers from an issue known as \textit{overoptimization} or \textit{reward hacking}~\citep{stiennon2020learning,gao2023scaling,coste2024reward}, where policy optimization improves the proxy reward model while actually degrading performance with respect to the true human preference. To address this, several lines of research have emerged, focusing either on improving the Reinforcement Learning (RL) training process via constrained policy optimization \citep{moskovitz2024confronting,zhang2024overcoming,liu2024provably}, or on designing more robust reward models that provide a reliable learning signal for RL training \citep{coste2024reward,yang2024regularizing,liu2025rrm}. Our work focuses on the latter direction.

One of the reasons for reward hacking in RLHF is that a reward model trained on limited human-preference data is inevitably an imperfect surrogate for the true underlying human preference \citep{gao2023scaling}. A natural solution is to construct a more robust reward signal by combining multiple reward models, therefore reducing reliance on any single imperfect proxy. Along this line, \citet{coste2024reward} average the predictions of multiple reward models to reduce the noise of individual models. Nevertheless, the effectiveness of such ensemble-based methods is constrained by the fact that reward models trained on similar data and objectives often share systematic blind spots, causing them to ``herd'' toward the same erroneous judgments \citep{eisenstein2024helping}. This suggests that robustness requires not merely aggregating multiple reward models, but explicitly promoting diversity among them and learning how to combine their complementary strengths.

Motivated by these limitations, we propose \textbf{Reward Model Boosting (\methodname)}, where we construct a more reliable reward signal with advanced boosting methods. 
First, we train a group of reward models jointly with a diversity-promoting penalty; in particular, we apply the Hilbert--Schmidt Independence Criterion \citep[HSIC;][]{gretton2005measuring,gretton2007kernel}, which reduces the correlation among models and promotes complementary error profiles. 
Then, we train a lightweight aggregator, following the principle of boosting, to combine the reward models effectively. Specifically, we learn a decision tree-based aggregator \citep{chen2016xgboost} that combines the reward models' outputs and iteratively minimizes the aggregated reward prediction error. The boosted decision-tree aggregator exploits reward models' complementary errors, yielding a more accurate and robust reward signal.  Our approach offers two key advantages. (1) \emph{Stronger aggregation}. Our \methodname{}\ method learns to correct one reward model's errors using the strengths of others during boosting training, thus producing more accurate predictions than simple averaging (shown in \S\ref{sec:indepth_analysis}). (2) \emph{Robustness to text perturbation.} By using a boosted decision-tree aggregator, \methodname{} induces a discretized mapping from model outputs to final scores, effectively smoothing spurious fluctuations caused by training noise (shown in \S\ref{sec:robust_analysis}).

We conduct extensive experiments to validate the effectiveness of our approach. First, we demonstrate that \methodname{} significantly improves reward prediction accuracy on both in-distribution and out-of-distribution (OOD) datasets, indicating strong generalization ability (\S\ref{sec:reward_performance}). Second, we conduct Best-of-$N$ (BoN) sampling and PPO training experiments, showing that our \methodname{} effectively mitigates the reward hacking issue  (\S\ref{sec:bon_setting}). Third, our in-depth analysis verifies that the HSIC diversity-promoting regularizer diversifies the reward models, which is essential in our reward model boosting framework (\S\ref{sec:indepth_analysis}). In addition, we perform text perturbation as a reward model attack, which further demonstrates the robustness of our \methodname{} (\S\ref{sec:robust_analysis}). 

\section{Method}

In this section, we present our \methodname{} in detail. We first present the task formulation in \S\ref{sec:task_formulation}, and then introduce our boosting method in \S\ref{sec:RMB}. Finally, we discuss diverse reward modeling in \S\ref{sec:HSIC}, which is important for performance gain.

\subsection{Task Formulation}\label{sec:task_formulation}

Let a dataset $\mathcal D$ contain tuples of $(x, y_w, y_l)$, where $x$ is an input text, and $y_w$ and $y_l$ are two responses to $x$. A human has indicated a preference of $y_w$ over $y_l$ ($w$ for win and $l$ for loss).

Our \methodname{} involves a set of reward models ${r_1, r_2, ..., r_N}$, where each model $r_i$ maps an input--response pair $(x,y)$ to a scalar reward score. The reward scores form a vector:
\begin{equation}
\mathbf R(x,y) = [r_1(x,y), r_2(x,y), \cdots, r_N(x,y)].
\end{equation}
Typically, a reward model is trained to assign a higher score to a preferred response $y_w$ than to a disfavored one $y_l$ by minimizing a pairwise comparison logistic loss~\citep{ouyang2022training, gao2023scaling, coste2024reward, yang2024regularizing, liu2025rrm}. We further incorporate a diversity penalty, deferred to \S\ref{sec:HSIC}.

Our boosting objective is to learn an aggregator function $F$ combining this reward vector into a scalar reward score, which is used to evaluate the quality of a response $y$ given its input $x$ for RLHF.

\subsection{Boosting the Reward Models}\label{sec:RMB}
We propose to learn the aggregator function $F$ by adapting the gradient boosting framework, specifically following the principle of XGBoost \citep{chen2016xgboost}. The key idea is to construct $F_K$ as an ensemble of $K$ regression decision trees $(f_1, f_2, \cdots, f_K)$, where each tree is trained on the residuals (the errors) from the combined predictions of all preceding trees. This sequential, additive modeling allows the aggregator to capture sophisticated, nonlinear interactions among the base reward model outputs.

Formally, the aggregator at iteration $k$ is expressed as:
\begin{equation}
F_k(\mathbf{R}(x,y)) = \sum_{t=1}^{k} f_t(\mathbf{R}(x,y)),
\end{equation}
where each $f_t$ is a regression tree that maps the reward vector $\mathbf{R}(x,y)$ to a scalar value. This is accomplished by traversing from the root to a leaf node, guided by internal nodes that contain a split condition based on a feature (an element in $\mathbf{R}(x,y)$) and a threshold (e.g., if $\mathbf{R}(x,y)[1] > 0.5$, go left; otherwise, go right). The final prediction is a scalar score held by a leaf node at the end of the path. The learnable parameters of a regression tree are its internal node split conditions and the scores held by the leaf nodes.

\textbf{Training objective.}
Given a dataset $\mathcal{D}$ of pairwise preference annotations, where each training example is a tuple $(x, y_w, y_l)$, we aim to learn a reward function, modeled by $F_K$, that assigns a higher score to $(x,y_w)$ than to $(x,y_l)$.  Following boosting, trees are added iteratively so that at iteration $k$ we fit $f_k$ to reduce the residual errors left by $F_{k-1}$. The training objective at iteration $k$ is 
\begin{align}\label{eq:xgb_obj}
\mathcal{L}^{(k)}
&= \sum_{(x,y_w,y_l)\in\mathcal{D}}
L\Bigl(
F_k(\mathbf{R}(x,y_w)) - F_k(\mathbf{R}(x,y_l))
\Bigr)
+ \Omega(f_k)
\\
&= \sum_{(x,y_w,y_l)\in\mathcal{D}}
L\biggl(
\underbrace{F_{k-1}(\mathbf{R}(x,y_w)) - F_{k-1}(\mathbf{R}(x,y_l))}_{\text{residual errors}}
 +\,f_k(\mathbf{R}(x,y_w)) - f_k(\mathbf{R}(x,y_l))
\biggr)
+ \Omega(f_k).
\end{align}

The data-fitting term in Eqn.~(\ref{eq:xgb_obj}) uses the pairwise logistic loss:
\begin{equation}\label{eq:rank_obj}
\begin{split}
&L\!\left(F_k(\mathbf{R}(x,y_w))-\,F_k(\mathbf{R}(x,y_l))\right) 
\quad = -\log\!\left(\sigma\!\left(F_k(\mathbf{R}(x,y_w)) - F_k(\mathbf{R}(x,y_l))\right)\right),
\end{split}
\end{equation}
where $\sigma(z)=\tfrac{1}{1+e^{-z}}$. This encourages the preferred response $y_w$ to receive a higher score than~$y_l$. The regularizer $\Omega(f_k) = \gamma T_k + \frac{\lambda}{2} \sum_{j=1}^{T_k} w_{kj}^2
$ penalizes overly complicated trees and large leaf values, where $T_k$ is the number of leaves in tree $f_k$, and $w_{kj}$ is the value of leaf $j$, tree $k$. The coefficients $\gamma$ and $\lambda$ control the strength of the penalty.

\textbf{Learning the decision trees.}
The structure is learned through a greedy algorithm, starting from a single root node and iteratively adding branches. At each node, the algorithm searches for an optimal split condition (in the form of feature thresholding).  The quality of a potential split is measured by the reduction of the objective function in Eqn.~(\ref{eq:xgb_obj}),  referred to as a gain:
\begin{align} \label{eq:gain}
\text{Gain} = \mathcal{L}^{(k)}_{\text{before split}} - \mathcal{L}^{(k)}_{\text{after split}}. \end{align}

To compute the gain in a tractable way, we follow \citet{chen2016xgboost} and use a second-order Taylor expansion to approximate the objective function~(\ref{eq:rank_obj}). Details are provided in Appendix~\ref{apd:XGB_optimization}.

The process of adding tree branches stops when the best candidate gain falls below a threshold; once the tree structure is fixed, each leaf value is set by a closed form that minimizes the second-order approximation of Eqn.~(\ref{eq:xgb_obj}) with the $\ell_2$ penalty in $\Omega(\cdot)$. Likewise, we stop adding trees for boosting when the objective no longer improves with an additional tree. 

Overall, our \methodname{} approach iteratively minimizes the objective by discovering new tree structures, each focusing on data samples that are currently misranked. In this way, it progressively refines $F_M$ into an accurate reward model aggregator.

\subsection{Learning Diverse Reward Models for Boosting}\label{sec:HSIC}

Boosting works best when base reward models make accurate and complementary predictions. When these reward models' scores are highly correlated, they cannot contribute to loss reduction during the boosting process \citep{geng2007feature,pan2009feature,tolocsi2011classification,yeh2022graph,lyzhin2023tricks}.

A na\"ive approach to encourage diversity of reward model learning is to introduce stochasticity (e.g., different random seeds) in training \citep{coste2024reward,eisenstein2024helping}. However, models trained on the same data tend to converge to similar functions even with different random seeds \citep{kornblith2019similarity, chizat2019lazy}. We therefore explicitly introduce a diversity-promoting regularization term, in addition to the conventional reward model training objective. 

\textbf{Reward model learning.} 
Following \cite{bradley1952rank}, the main objective of reward model learning is to distinguish between a preferred response $y_w$ and a less preferred one $y_l$: 
\begin{equation}\label{eq:reward_obj}
\mathcal{L}_{\text{reward}} (\theta) = -\mathbb{E}_{(x, y_w, y_l) \sim \mathcal D} \left[\log \left( \sigma \left( r_\theta(x, y_w) - r_\theta(x, y_l) \right) \right) \right],
\end{equation}
where $r_\theta(x, y)$ represents the reward score for input $x$ and its corresponding response $y$ with model parameters~$\theta$, and 
$\sigma (\cdot)$ is the sigmoid function. 

\textbf{HSIC diversity regularization.} We adopt the Hilbert--Schmidt Independence Criterion \citep[HSIC;][]{gretton2005measuring,gretton2007kernel} to quantify and penalize statistical dependence between reward models. A key motivation is that redundancy among reward models is often \emph{nonlinear}: two models may exhibit weak global correlation yet rely on the same shortcut features and thus fail on the same prompt subpopulations. As a kernel-based dependence measure, HSIC is sensitive to a broad class of nonlinear relationships, making it well-suited for encouraging genuinely complementary reward signals. Moreover, HSIC yields a direct dependence penalty that is easy to compute on minibatches and integrates naturally as a differentiable regularizer during training. 

Importantly, we do not apply HSIC to raw reward scores $r_\theta(x,y)$, since reward scales and offsets are not semantically identified and directly decorrelating raw scores can conflict with the primary preference-learning objective in Eqn.~(\ref{eq:reward_obj}). Instead, we regularize dependence at the level of \emph{preference margins}, which preserves invariance to per-model affine transformations while targeting diversity in the decision-relevant quantity that drives pairwise comparisons.

Let $r_{\theta_1}, \cdots, r_{\theta_N}$ be $N$ reward models and $\{(x^{(i)}, y_w^{(i)}, y_l^{(i)})\}_{i=1}^b$ be a batch of samples. We introduce a margin vector $\mathbf m_k$ for the $k$-th reward model:
\begin{align}\label{eq:margin_vec}
\mathbf{m}_k \;=\; \big[\, m_k^{(1)},\cdots,m_k^{(b)} \,\big]^\top
\quad\text{with}\quad 
m_k^{(i)} \;=\; r_{\theta_k}(x^{(i)}, y_w^{(i)})\;-\; r_{\theta_k}(x^{(i)}, y_l^{(i)}).
\end{align}
These margins quantify how strongly model $k$ prefers $y_w^{(i)}$ over $y_l^{(i)}$ for input $x^{(i)}$. If two models yield nearly identical margin vectors on a batch, they are learning essentially the same notion of preference strength, i.e., collapsing to a shared margin distribution. Promoting diversity based on the margin vectors helps to learn diverse reward models in a way that they agree on sign, but differ in agreement strength.

Specifically, HSIC uses a kernel function to create a unique representation for each reward model in a high-dimensional space~\citep{gretton2005measuring,ullah2024conventional,birjais2024training}. HSIC checks whether the pattern of one reward model aligns with another in the high-dimensional space based on pairs of margin vectors, given by $\operatorname{HSIC}_b(\mathbf m_i, \mathbf m_j)$. Details of this HSIC regularizer computation are provided in Appendix~\ref{apd:hsci_details}.

In summary, we train all $N$ reward models jointly with the following objective:
\vspace*{-9pt}
\begin{equation}
\label{eq:joint_loss_margins}
\begin{split}
\mathcal{L}_{\text{total}}(\theta_1,\cdots,\theta_N)
&= \sum_{k=1}^N \mathcal{L}_{\text{reward}}(\theta_k) +
 \lambda_{\text{HSIC}}
\sum_{1 \le i < j \le N} 
\operatorname{HSIC}_b\!\big(\mathbf{m}_i,\mathbf{m}_j\big),
\end{split}
\end{equation}
where $\lambda_{\text{HSIC}}>0$ controls the strength of the diversity regularization. Minimizing this joint objective (\ref{eq:joint_loss_margins}) yields an ensemble of reward models that are not only individually accurate but also exhibit diversity in preference prediction. 

\section{Experiments}\label{sec:exps} In this section, we present the empirical evaluation of our proposed \methodname. We begin by describing the datasets, models, and competing approaches, followed by our main results and in-depth analyses.

\subsection{Experimental Settings} \textbf{Datasets.} We evaluate our approach on four human preference datasets commonly used in prior work \citep{zhang2025bradley,mandal2025distributionally}. \textbf{Unified-Feedback}\footnote{Unified-Feedback: \href{https://huggingface.co/datasets/llm-blender/Unified-Feedback}{https://huggingface.co/datasets/llm-blender/Unified-Feedback}}: One of the largest publicly available collections of pairwise human feedback. 
\textbf{HHH-Alignment} \citep{askell2021general}: A preference dataset constructed to evaluate language models on helpfulness, honesty, and harmlessness. 
\textbf{MT-Bench Human Judgments} \citep{zheng2023judging}: A collection of human preferences for model responses to questions from the MT-Bench benchmark. 
\textbf{RewardBench} \citep{lambert2024rewardbench}: A benchmark created to evaluate the capabilities and safety alignment of reward models.

We use the Unified-Feedback dataset for training and in-distribution evaluation; we treat HHH-Alignment, MT-Bench, and RewardBench as out-of-distribution (OOD) evaluation sets, as their input and response distributions differ significantly from those of the training data. \revision{Notice that HHH-Alignment, MT-Bench, and RewardBench contain only 221, 5,755, and 2,985 evaluation samples, respectively, and do not have training samples. Thus, we use them exclusively for OOD evaluation, following previous work~\citep{yang2024regularizing}.}

\input{40k_main}

\textbf{Models.}\label{sec:models} { }Following the setup of \citet{yang2024regularizing}, we use Gemma-2B-IT \citep{team2024gemma} as the backbone for both reward model training and policy optimization;  we use a publicly available Mistral-7B-Instruct-v0.2 model\footnote{Mistral-7B-Instruct-v0.2 model: \href{https://huggingface.co/Ray2333/reward-model-Mistral-7B-instruct-Unified-Feedback}{https://huggingface.co/Ray2333/reward-model-Mistral-7B-instruct-Unified-Feedback} } that was fine-tuned on the Unified-Feedback dataset as the gold reward model serving for a synthetic setup to detect the reward hacking issue. 

\textbf{Competing approaches.} { }We compare the performance of \methodname{} against six competing approaches: 
\textbf{Baseline Reward Model:} It is a standard model trained with the classic reward objective defined in Eqn.~(\ref{eq:reward_obj}). 
\textbf{Frozen Reward Model:} Here, the reward model reuses the base language model's weights, except that a two-layer neural classifier is fine-tuned.
\textbf{Margin Loss:} An additional margin is incorporated into the reward loss, as shown by Eqn.~(\ref{eq:reward_loss_margin}) in Appendix~\ref{appendix:implement_details} \citep{touvron2023llama,wang2024secrets}. 
\textbf{Label Smoothing:} A regularization technique that penalizes overconfident model outputs to mitigate overfitting \citep{wang2024secrets}. 
\textbf{GRM:} The GRM method employs a regularization term of the text generation objective during reward model learning \citep{yang2024regularizing}. 
\textbf{Average Ensemble:} A method that averages the outputs of a group of reward models learned in the same way as the baseline reward model with different random seeds as the final reward \citep{eisenstein2024helping}. Unless stated otherwise, we average outputs of three reward models as an ensemble.

\revision{
All competing approaches and RMB are evaluated under the same controlled experimental setup, including identical data splits, preprocessing, backbone models, and training frameworks.} For implementation details of all approaches, please refer to Appendix~\ref{appendix:implement_details}.

\subsection{Reward Model Performance}\label{sec:reward_model_performance}
We evaluate the effectiveness of our approach through two main experiments: (1) reward model performance on human-preference data (i.e., whether the learned reward model can rank human preferred responses with a higher score or not), and (2) generation performance of the policy learned by RLHF based on different reward models \citep{ouyang2022training}.

\textbf{Preference prediction performance.} \label{sec:reward_performance}
Following the setup of \citet{yang2024regularizing}, we learn reward models using Unified-Feedback with two training sizes: 40K samples (Tab.~\ref{tab:reward_performance_40k} in the main paper) and 400K (Tab.~\ref{tab:reward_performance_400k} in Appendix~\ref{sec:efficiency}). We compute the accuracy that a reward model correctly ranks the win/lose responses (i.e., assigning a higher reward to the win response). 
We use the data processing code released by \cite{yang2024regularizing} to ensure a fair comparison.

As seen from the tables, recent methods of learning reward models---such as Label Smooth~\citep{wang2024secrets}, Margin~\citep{touvron2023llama}, GRM~\citep{yang2024regularizing}, Avg Ensemble~\citep{coste2024reward}---bring marginal improvement compared with the baseline approach~\citep{ouyang2022training}. By contrast, our \methodname{} approach consistently and substantially outperforms all competing methods in both in-distribution and OOD setups. In particular, our model surpasses the classic average ensemble \citep{coste2024reward} when the number of ensemble individuals is controlled ($N=3$ and $N=5$ in our experiments).

\textbf{LLM-as-judge evaluation of policies learned via RLHF.}
\input{win_rate_pairwise}
{ }We use an LLM judge to evaluate the policies (i.e., language models) learned by RLHF. Specifically, we use Gemma-2B-IT as a base policy model and fine-tune it by the PPO algorithm using the reward models learned in the previous setup. Then, we prompt DeepSeek-R1 \citep{deepseekai2025deepseekr1incentivizingreasoningcapability} to judge the overall quality of the responses for the inputs in the Unified-Feedback hold-out test set. We select three competitors in this evaluation due to the limitation of budget: Baseline RM as a baseline method, GRM, and Avg Ensemble as strong competitors. To mitigate ID and positional biases \citep{zheng2023large,shen2023large}, we conduct pairwise comparison in a way that each pair is queried four times by swapping the candidate order and their IDs (i.e., ``A'' and ``B''). In addition, we also conduct a pointwise comparison, where we obtain an evaluation score ranging from 1 to 10; then, we compute a win/tie/loss rate according to the score comparison. LLM evaluation prompt templates are provided in Appendix~\ref{apd:llm-eval}.

Tab.~\ref{tab:llm_eval} reports the results of LLM-as-judge. As seen, \methodname{} yields an LLM policy that attains the highest win rate. This is consistent against all competitors and in both pairwise and pointwise settings. These findings, along with the results in the preference prediction experiment, convincingly demonstrate the effectiveness of our \methodname{} method for RLHF.

\subsection{Evaluation of the Mitigation of Reward Hacking}

We analyze the reward hacking by Best-of-N (BoN) sampling and Proximal Policy Optimization (PPO) experiments, following prior work \citep{gao2023scaling}. This involves a synthetic data setup, where we use Mistral-7B-Instruct-v0.2 as the gold reward to provide preference annotations for training a reward model (called a proxy reward). 
\input{bon}
\textbf{BoN.}\label{sec:bon_setting}
{ } BoN sampling draws $n$ candidate responses from a general language model (i.e., Gemma-2B-IT) and picks the one with the highest score under a proxy reward model. As $n$ increases, if the chosen output’s proxy score rises while its gold reward score plateaus or even declines, then it shows that this proxy reward model suffers from reward hacking. BoN sampling has been widely adopted as an experiment setup to illustrate the overoptimization issue \citep{gao2023scaling,yang2024regularizing,liu2025rrm}.

Our experiment adopts the setup in  \citet{yang2024regularizing}. We first draw a subset containing $20\text{K}$ samples from Unified-Feedback and annotate preference labels (i.e., which sample is preferred) based on the gold reward to train a proxy reward model. For each input in the test set, we perform BoN sampling (i.e., drawing $n$ candidates and ranking them with the proxy model). The top-ranked response is selected and then scored by the gold reward.  We report both the average proxy score and the average gold score across the test inputs. We vary the selection budget via the BoN KL divergence from $0$ to $5$,  corresponding to the number of responses $n$ ranging from $1$ to $405$ per input. The KL divergence is given by
$\mathrm{KL}_{\text{BoN}} = \log n - \frac{n-1}{n}$, which quantifies how $n$-category uniform distribution deviates from the one-hot distribution of a sample.

Figs.~\ref{fig:bon-a} and~\ref{fig:bon-b} show the performance of our \methodname{} approach and competing approaches in the BoN sampling experiment. We see that, for competing approaches, the gold reward score increases much less than the proxy reward score as $\mathrm{KL}$ grows, indicating reward hacking exists. By contrast, \methodname{} exhibits a clear, monotonic increase in the gold reward score with larger $\mathrm{KL}$, demonstrating its advantage in mitigating reward hacking.

It is noticed that human-preference data typically contains $\sim\!20$–$30\%$ label noise \citep{wang2024secrets}, which may degrade reward model generalization \citep{rame2024warm,liang2024robust} and hinder policy learning \citep{ye2024corruption,mandal2025corruption}. To assess robustness under label noise, we randomly choose $25\%$ of the data samples and flip their gold reward-induced preference labels. 

The results in Figs.~\ref{fig:bon-c} and~\ref{fig:bon-d} show that, with label noise, the baseline suffers from significant reward hacking, as its gold reward score decreases. GRM and Avg Ensemble demonstrate some capacity of mitigating reward hacking, but the gold reward score achieved still plateaus as their proxy reward scores continue to increase. 
By contrast, \methodname{} achieves the highest gold reward score, which increases robustly. This indicates that our approach can reliably estimate the response quality even when trained on noisy preference labels.

\textbf{PPO.} \label{sec:ppo_performance}
{ } PPO~\citep{schulman2017proximal} is a classic RL algorithm designed to learn a policy maximizing expected rewards. It has been widely adopted in RLHF to learn a language model policy maximizing the preference score provided by reward models~\citep{ouyang2022training,achiam2023gpt,liu2024provably} and also serves as a standard testbed for studying the reward hacking issue~\citep {gao2023scaling,moskovitz2024confronting,bukharin2025adversarial}.
\input{ppo}
Specifically, we adopt the Gemma-2B-IT model as an initial policy, and use the PPO algorithm to optimize this policy based on the proxy reward model (which is learned by gold reward-induced preference labels). 
The PPO-finetuned policy is then used to generate responses on the inputs from the test set of Unified-Feedback. We track the changes of both proxy and gold reward scores of the generated response throughout PPO training to assess whether improvement under the proxy reward score translates to that under the gold reward score.

As seen from Figs.~\ref{fig:ppo-a} and~\ref{fig:ppo-b}, all methods increase the proxy reward score at the beginning of training; however, competing approaches exhibit noisier trajectories with a noticeable mid-training drop, whereas \methodname{} shows smoother and more stable improvements. For the gold score, competing approaches improve briefly only in the first few steps and collapse later. On the contrary, the curve of the gold reward in our approach is more reflected by the proxy reward curve. The results suggest that our approach is able to substantially mitigate reward hacking.

As in the BoN experiment, we also have a setup with $25\%$ noise added to preference labels in the PPO experiment, and the learning curves are shown in
Figs.~\ref{fig:ppo-c} and~\ref{fig:ppo-d}. As expected, the gold reward score is more fluctuated due to the added label noise. But still, our \methodname{} maintains the highest and least decreasing gold score, demonstrating that our approach is more robust than competing methods with noisy preference labels.

\subsection{Analysis of the Components of \methodname{} }\label{sec:indepth_analysis}

\input{ablation}

We perform an in-depth analysis of two main components of our approach, namely, the boosting method and the HSIC diversity-promoting regularization. We follow most of the setups in \S\ref{sec:reward_performance}, except that we consider five\footnote{
In the main experiment, we try different numbers of individuals ($N=3, 5, 8, 10$) for our \methodname{}, but compare our approach with Avg Ensemble in the setting of $N=3$ as we quote results from previous work for Tab.~\ref{tab:reward_performance_40k}. Here, we consider five individual RMs to have a more reliable correlation analysis.

} individual reward models (RMs) by either varying random seeds or using our HSIC diversity-promoting regularizer.

As shown in Tab.~\ref{tab:ablation_table}, Average Ensemble and \methodname{} achieve similar performance in the random seed setup, whereas with HSIC-induced diversity, our boosting considerably outperforms Average Ensemble. This shows that different random seeds yield similar reward models, thus making ensembles less effective. By contrast, our HSIC encourages complementary error profiles that a boosting aggregator can leverage to achieve substantially higher performance.

The correlation among RMs can be further viewed from the heatmaps in Fig.~\ref{fig:correlation}. For each pair of RMs, we measure the ratio of pairwise-preference agreement on the Unified-Feedback test set. RMs learned under random seed diversity exhibit very high pairwise agreement (most pairs $>\!80\%$), indicating strong correlations; HSIC-trained RMs are less correlated (typically $50$--$80\%$), providing the diversity that boosting needs.

\input{feature_correlation}
We further examine the contribution of each RM during the boosting process. This is measured by \emph{Total Gain}~\citep{chen2016xgboost}, i.e., the sum of gains over all splits (see Eqn.~(\ref{eq:gain})) that each RM obtains in the decision trees in Tab.~\ref{tab:diversity_gain}. We see that, under the random seeds setup, Total Gain is heavily skewed (e.g., a single RM contributes $70.13\%$). This is because such RMs tend to be redundant and thus are ineffective when the boosting process minimizes residual errors.

\input{feature_weights}
Under the HSIC setup, the Total Gain is distributed more evenly. RM~0, RM~4, and RM~1 each contribute $25$--$30\%$, and they are the strong individual RMs (seen in Tab.~\ref{tab:ablation_table}).
This shows that boosting emphasizes the best RMs while using the remaining RMs to correct residual errors. Overall, our boosting approach surpasses every individual RM.

\textbf{Discussion.}
The low correlation of individuals is shown to be important for boosting algorithms in commercial ranking systems, such as search and recommendation~\citep{covington2016deep,wang2022learning,xi2023device}, for which researchers perform heavy feature engineering and propose heterogeneous signals (e.g., semantic relevance, like count, and comment count) to be considered during boosting~\citep{ni2021prioritizing,malay2023airbnb,du2024disentangled}. In our setting, HSIC serves as an auxiliary objective that encourages low pairwise correlation across RMs, which is analogous to the feature engineering process of traditional boosting systems.

\subsection{Analysis of Robustness under Attacks}\label{sec:robust_analysis}

We evaluate RM robustness under small, meaning-preserving text perturbations. This is accomplished by comparing how an RM ranks a set of response candidates for the same input, with and without perturbation. Specifically, we sample 400 candidate responses for each input in the test set of Unified-Feedback using a pretrained Gemma-2B-IT. We first compute reward scores on the original texts and obtain a ranking in descending order for each input.  We then apply a keyboard-neighbor typo attack to the text by randomly replacing characters with adjacent keys on the QWERTY layout~\citep{belinkov2018synthetic,pruthi2019combating}, and compute a new rank. To study sensitivity, we vary the perturbation budget (number of character edits per example) in \{1, 3, 5, 8, 10\}.

We measure the similarity of the two rankings by three metrics: (1) \textbf{Pairwise accuracy}: the ratio of correctly ordered pairs among all pairs, (2) \textbf{Footrule similarity}: one minus normalized Spearman’s footrule distance, which sums the absolute differences between the ranks of each item in the two rankings~\citep{diaconis1977spearman}, and (3) \textbf{Normalized Discounted Cumulative Gain (nDCG)}: a metric that measures the usefulness of items based on their position in the rank~\citep{jarvelin2002cumulated}, where we treat the original order as the ground truth.
All metrics are in the range of $[0,1]$, and the higher, the better. 

Figure~\ref{fig:robustness_plt} shows the results of our \methodname{} ($N=3$), along with the Baseline RM, Avg Ensemble ($N=3$), and GRM. As seen, our \methodname{} attains the highest scores on all three metrics, indicating that our \methodname{} is more robust than competing approaches under small, meaning-preserving attacks. 
\input{robustness}

\textbf{Discussion.} Decision-tree-based aggregators are known to be robust in ranking tasks~\citep{joachims2017unbiased,wang2018position,haldar2019applying, wu2022neural}, which partially explains that our \methodname{} can maintain high performance during attacks. It is interesting to notice that a Transformer~\citep{vaswani2017attention}-based reward model maps discrete text to continuous reward values, and a small change of the input text may lead to large changes of the output, especially when the training set is small~\citep{ wu2025rewordbench}; such a phenomenon is observed in other tasks as well~\citep{ebrahimi2018hotflip, jin2020bert}. Our \methodname{} performs aggregation of individual RMs by only predicting one of its leaf values, which are a finite discrete set. This reduces spurious fluctuations caused by training noise, making our model more robust under attacks.

\subsection{Efficiency Analysis}
Our \methodname{} method introduces a lightweight aggregator on top of a set of base RMs. To verify its efficiency, we evaluate its impact on inference latency. Specifically, we decompose the inference process into two stages: (1) the \textbf{base RM pass}, which involves a forward pass through each of the three RMs to generate individual scores, and (2) the \textbf{aggregation step}, where individual RM scores are combined into a final prediction. We measure the time spent in each stage for both our \methodname{} aggregator and Average Ensemble~\citep{coste2024reward}. Metrics include
\begin{compactitem}
    \item \textbf{Breakdown latency} (s/batch): The wall-clock time required for a given stage. We report the mean, median (p50), 90th (p90), and 99th (p99) percentile latencies across all batches.
    \item \textbf{End-to-end latency} (s/batch): The overall latency including RMs and the aggregator, as well as other code components (such as IO).
\end{compactitem}

The results are summarized in Tab.~\ref{tab:efficiency}. As seen, the main computation is devoted to the base RM pass, with a mean latency of $19.791$ seconds per batch. In comparison, the Average Ensemble aggregator adds a mean overhead of only $5.92\times 10^{-5}\,\mathrm{s}$ ($\approx 0.06\,\mathrm{ms}$). Similarly, our \methodname{} aggregator is also efficient as the overhead is magnitudes lower than the base RM pass, as the aggregator vs.~end-to-end ratio is only $0.000088$. The results show that our RMB is a lightweight aggregator. It introduces a negligible latency to the reward model inference. Thus, it also introduces a negligible latency to RLHF training when the rest of the algorithm and implementation are controlled.

\input{efficiency}

\subsection{Results in Computational Budget-Constrained Scenario}
\input{param_control}
We further examine whether the gains of \methodname{} hold when the computational budget is limited. In this setting, we compare boosting with $N$ reward models against training a single larger RM that has a similar or larger number of parameters. We perform full-parameter fine-tuning for one epoch using the open-source dataset mixture of HH-RLHF, SHP, UltraFeedback, and Summarization.\footnote{preference-700K: \href{https://huggingface.co/datasets/hendrydong/preference_700K}{https://huggingface.co/datasets/hendrydong/preference-700K}} This mixture was introduced by~\citet{dong2024rlhf}, who train FsfairX-LLaMA3-RM-8B on it. To ensure that our results are directly comparable, we use the same training data and evaluate on RewardBench as the test set.

We also include stronger competing reward models whose parameter counts are much larger than ours, including a 34B reward model\footnote{Starling-RM-34B: \href{https://huggingface.co/Nexusflow/Starling-RM-34B}{https://huggingface.co/Nexusflow/Starling-RM-34B}} and GPT-4/GPT-4o used as judges. Tab.~\ref{tab:reward_model_comparison} shows that \methodname{} performs well in this budget-constrained scenario: it outperforms the 34B reward model and also surpasses GPT-4/GPT-4o as judges. In addition, \methodname{} yields a clear improvement over the 8B baseline, raising the score from 84.4 to 87.8, while using an ensemble with fewer total parameters and the same training data. Overall, these results confirm that \methodname{} remains effective when computational resources are constrained.

\section{Related Work}

The challenge of reward hacking in RLHF was empirically established by a number of studies~\citep{ouyang2022training, gao2023scaling}. There are generally two directions to mitigate the reward hacking issue: (i) improving the RL training process under an imperfect reward signal, and (ii) learning a better reward model to provide a reliable training signal for RL.

\textbf{Optimizing the RL process.} One line of work mitigates reward hacking by refining the RL optimization process under imperfect reward signals. For example, researchers update the policy conservatively by designing pessimistic reward signals given an imperfect RM~\citep{zhai2023uncertainty, zhang2024overcoming, yan2024reward, mandal2025distributionally}. Another strategy is to maintain a pretrained policy's text generation ability by regularizing the RL objective~\citep{ouyang2022training,liu2024provably,laidlaw2025correlated}, or averaging policies' parameters~\citep{lin2024mitigating}. Researchers also constrain the policy optimization by restricting the exploration region~\citep{moskovitz2024confronting, dai2025mitigating, mandal2025corruption}. 

\textbf{Improving reward models.} Our work falls into this category, which combats reward hacking by strengthening reward models. Building an ensemble of multiple RMs can stabilize noisy reward signals; for example, \citet{coste2024reward} average RM outputs and \cite{rame2024warm} average RM parameters. However, simple averaging cannot remove the bias shared among individual RMs~\citep{eisenstein2024helping}. To address this, our RMB learns an aggregator based on the principle of boosting~\citep{chen2016xgboost}, which reduces residual reward error in a complementary way.

Most other RM-centric studies focus on improving an individual reward model, including designing better RM training objectives~\citep{yang2024regularizing,wang2024interpretable}, optimizing RM architectures~\citep {miao2024inform,chen2024odin}, and improving training data quality~\citep{zhu2024iterative, wolf2025reward, haider2025framework, liu2025rrm,bukharin2025adversarial}. These approaches focus on a single RM, which can be potentially combined with our boosting approach for further performance improvement. 

\revision{
\textbf{Multi-objective alignment.}
Another related line of work studies alignment with multiple preference objectives, where optimizing one objective may interfere with others. Representative approaches address such conflicts during policy optimization by adaptively combining objective-specific gradients \citep{li2025gradient,chen2026rewardfree}, while recent work also studies conflicts arising from preference data \citep{xu2026understanding}. This line of research mainly concerns how multiple objectives can be jointly optimized without undesirable trade-offs or collapse. RMB instead focuses on an earlier stage of the RLHF pipeline: constructing a reliable reward signal before policy optimization. Rather than assigning individual RMs to predefined objectives, RMB applies diversity regularization to encourage different RMs trained on the same preference data to capture latent complementary preference patterns, and then uses boosting to aggregate them into a robust reward signal. Therefore, the two directions largely address orthogonal components of alignment and are difficult to compare directly. An interesting future direction is to investigate whether the complementary patterns learned by RMB correspond to interpretable preference dimensions, such as helpfulness, harmlessness, or honesty, and whether such reward representations can further support downstream multi-objective alignment. 

In short, RLHF is a promising research direction, and various methods have been proposed to improve RLHF at different stages. Due to the scope of this paper, our empirical comparisons focus on methods that directly compete with RMB at the reward-modeling stage.
}

\section{Conclusion}
To address the challenge of reward hacking in RLHF, we propose Reward Model Boosting (RMB). Our method learns a decision-tree aggregator over an ensemble of reward models to produce a more accurate and robust reward signal. We further adopt the HSIC diversity-promoting regularizer for reward model training. Our experimental results show that RMB enhances preference prediction accuracy across both in-distribution and out-of-distribution datasets. Moreover, our RMB effectively mitigates reward hacking and stays robust when countering text perturbation. Overall, RMB provides a reliable reward signal for training policies to align with human preferences.

\section{Limitations}
\label{sec:limitation}
\revision{
RMB relies on each reward model providing complementary preference signals. In our experiments, the marginal gain diminishes as the number of reward models increases, with performance largely saturating beyond a moderate aggregation size. One possible explanation is that the training data do not contain sufficiently rich and independent learning signals: as more reward models are added, they increasingly learn redundant preference patterns, limiting the benefits of diversity regularization and boosting. Since our experiments are constrained by the coverage and quality of currently available open-source preference datasets, we cannot fully evaluate RMB's scaling potential under richer and more heterogeneous feedback. Evaluating RMB on larger, higher-quality, and more diverse preference corpora is therefore an important direction for future work.

Another limitation is that the boosting stage may not preserve preference order if the RMs are added with a constant offset, due to the unidentifiability of the Bradley--Terry objective. However, this is less of a concern in our approach, because we first train RMs, based on which the boosting model is then trained. In this way, the boosting model can account for the offset of RMs and adapt accordingly. Nevertheless, we conduct an analysis on the sensitivity of our RMB under reward shifts in Appendix~\ref{app:shift_robustness}. We observe that injecting the constant into RMB results in only small changes to its pairwise rankings, indicating 
RMB’s rankings remain largely stable if RMs are shifted.

}
\section*{Acknowledgments}
We thank all reviewers and editors for their insightful feedback and constructive comments. This research
was supported in part by the Natural Sciences and Engineering Research Council of Canada (NSERC), the Amii Fellow Program, the Canada CIFAR AI Chair Program, and the Digital
Research Alliance of Canada.

\bibliography{main}
\bibliographystyle{tmlr}

\appendix
\section{RMB Details}
\subsection{Optimization of the Reward Model Boosting Objective}\label{apd:XGB_optimization}

This section provides a detailed derivation for learning the tree aggregator, $f_m$, at each boosting iteration $m$. The primary goal is to determine the optimal tree structure and leaf values that minimize the overall objective function defined in Eqn.~(\ref{eq:xgb_obj}). The optimization follows the principle of XGBoost~\citep{chen2016xgboost}.

\paragraph{Objective function approximation.}
The training objective at the $m$-th iteration, as presented in Eqn.~(\ref{eq:xgb_obj}) of the main text, can be rewritten as the following form:
\begin{equation}
\mathcal{L}^{(k)} = \sum_{(x,y_{w},y_{l})\in\mathcal{D}}  L(\Delta F_{k-1}(\mathbf{R}) + \Delta f_{k}(\mathbf{R})) + \Omega(f_{k}),
\end{equation}
where $\Delta F_{k-1}(\mathbf{R}) = F_{k-1}(\mathbf{R}(x,y_{w})) - F_{k-1}(\mathbf{R}(x,y_{l}))$ and $\Delta f_{k}(\mathbf{R}) = f_{k}(\mathbf{R}(x,y_{w})) - f_{k}(\mathbf{R}(x,y_{l}))$. $\Omega(f_k)$ is a regularizer of a tree to penalize overly complicated trees structure and large leaf values To facilitate efficient optimization, we approximate the loss function $L$ using a second-order Taylor expansion around the prediction from the previous iteration, $\hat{y}^{(k-1)} = \Delta F_{k-1}(\mathbf{R})$, converting the problem into minimizing a quadratic form.

Let $g_m$ and $h_m$ be the first and second-order gradients of the loss function with respect to the prediction for the $m$-th preference data point in the training dataset, denoted by 
\begin{align}
g_m &=\left. \frac{\partial L(y_m, \hat{y})}{\partial \hat{y}}\right|_{\hat{y} = \hat{y}^{(k-1)}_m} \\
h_m &= \left. \frac{\partial^2 L(y_m, \hat{y})}{\partial \hat{y}^2} \right|_{\hat{y} = \hat{y}^{(k-1)}_m}
\end{align}
The loss function for a single instance can thus be approximated as:
\begin{equation}
L(\hat{y}^{(k-1)}_m + \Delta f_{k}(\mathbf{R})) \approx L(\hat{y}^{(k-1)}_m) + g_m \Delta f_{k}(\mathbf{R}) + \frac{1}{2}h_m \Delta f_{k}^2(\mathbf{R}).
\end{equation}
After removing the constant term $L(\hat{y}^{(k-1)}_m)$, we simplify the objective function to:
\begin{equation}
\tilde{\mathcal{L}}^{(k)} \approx \sum_{m=1}^{M} \left[ g_m \Delta f_{k}(\mathbf{R}_m) + \frac{1}{2}h_m \Delta f_{k}^2(\mathbf{R}_m) \right] + \Omega(f_k).
\label{eq:approx_obj}
\end{equation}

\paragraph{Expressing the objective by the parameters of decision trees.}
A decision tree $f_k$ is defined by its structure $q$ and leaf values~$\mathbf {w}$. The structure $q: \mathbb R^d\rightarrow\{1,\cdots, T\}$ maps an input instance $\mathbf{R}\in\mathbb R^N$ to a specific leaf index in $1,\cdots, T$, and $w_j$ is the score assigned to the $j$th leaf node for $j=1,\cdots, T$.

To express the parameters explicitly in the objective in Eqn.~\ref{eq:approx_obj}, we regroup the sum over training instances to a sum over the disjoint sets of instances contained in each leaf. 
In particular, the term $\Delta f_{k}(\mathbf{R})$ can be expressed as the difference between the leaf values corresponding to the winning and losing responses: $\Delta f_{k}(\mathbf{R}) = w_{q(\mathbf{R}_{w})} - w_{q(\mathbf{R}_{l})}$. Let $I_j = \{k : q(\mathbf{R}_k) = j\}$ denote the set of instances partitioned into leaf $j$. By regrouping the summation in Eqn.~(\ref{eq:approx_obj}) over the leaves, we obtain:
\begin{equation}
\tilde{\mathcal{L}}^{(k)} = \sum_{j=1}^{T} \left[ \left(\sum_{m \in I_j} g_m\right) w_j + \frac{1}{2} \left(\sum_{m \in I_j} h_m\right) w_j^2 \right] + \gamma T + \frac{1}{2}\lambda \sum_{j=1}^{T} w_j^2
\end{equation}
where the last two terms are the regularizer, i.e., $\Omega(f_k)=\gamma T + \frac{1}{2}\lambda \sum_{j=1}^{T} w_j^2$. Let us define the sum of gradient statistics in each leaf as $G_j = \sum_{m \in I_j} g_m$ and $H_j = \sum_{m \in I_j} h_m$. The objective can be concisely written as:
\begin{equation}
\tilde{\mathcal{L}}^{(k)} = \sum_{j=1}^{T} \left[ G_j w_j + \frac{1}{2}(H_j + \lambda) w_j^2 \right] + \gamma T.
\label{eq:leaf_obj}
\end{equation}

\paragraph{Optimal leaf weight calculation.}
 
The objective function in Eqn.~(\ref{eq:leaf_obj}) is a sum of independent quadratic functions of each leaf weight $w_j$. To find the optimal value $w_j^*$, we take the partial derivative of $\tilde{\mathcal{L}}^{(k)}$ with respect to $w_j$ and set it to zero:
\begin{equation}
\frac{\partial \tilde{\mathcal{L}}^{(k)}}{\partial w_j} = G_j + (H_j + \lambda)w_j = 0,
\end{equation}
which in turn yields the optimal leaf weight:
\begin{equation}
w_j^* = -\frac{G_j}{H_j + \lambda}.
\label{eq:opt_weight}
\end{equation}

\paragraph{Decision-tree structure learning.}

Finding the globally optimal tree structure is difficult, as it would require evaluating an exponential number of possible tree structures. Instead, we adopt a greedy, top-down approach to build the tree by making the best possible split at each step, which is computationally feasible and effective in practice. 

We use \emph{gain} in Eqn.~(\ref{eq:gain}) to quantify the quality of a potential split. Specifically, we first substitute the optimal leaf weight $w_j^*$ in Eqn.~(\ref{eq:opt_weight}) back into the objective function in Eqn.~(\ref{eq:leaf_obj}). This yields  a score that evaluates the quality of a given tree structure $q$:
\begin{equation}
\tilde{\mathcal{L}}^{(m)}(q) = -\frac{1}{2} \sum_{j=1}^{T} \frac{G_j^2}{H_j + \lambda} + \gamma T.
\label{eq:tree_score}
\end{equation}
Consider a split that partitions a parent leaf into two child leaves, left and right. The gain of this split is defined as the reduction in the objective function:
\begin{equation}
\text{Gain} = \tilde{\mathcal{L}}_{\text{parent}} - (\tilde{\mathcal{L}}_{\text{left\_child}} + \tilde{\mathcal{L}}_{\text{right\_child}}).
\end{equation}
Combining Eqn.~(\ref{eq:tree_score}), we have
\begin{equation}
\text{Gain} = \frac{1}{2} \left[ \underbrace{\frac{G_L^2}{H_L + \lambda}}_{\text{Left Score}} + \underbrace{\frac{G_R^2}{H_R + \lambda}}_{\text{Right Score}} - \underbrace{\frac{G^2}{H + \lambda}}_{\text{Parent Score}} \right] - \underbrace{\gamma T}_{\text{Penalty}}.
\end{equation}
The algorithm enumerates all possible splits over all entries in $\mathbf{R}$ and selects the split that maximizes this gain. The process is repeated until a stopping criterion is met (i.e., reaching a maximum depth or observing no positive gain).

\subsection{Details of the HSIC Diversity-Promoting Regularizer} \label{apd:hsci_details}

In the \S\ref{sec:HSIC}, we aim to train a diverse ensemble of $N$ reward models $\{R_{\theta_n}\}_{n=1}^N$. Specifically, we apply the Hilbert--Schmidt Independence Criterion \citep[HSIC;][]{gretton2005measuring,gretton2007kernel} to the preference margins.

Consider a batch of $b$ preference pairs $D_b = \{(x_w^{(i)}, x_l^{(i)})\}_{i=1}^b$ and two reward models $R_{\theta_m}$ and $R_{\theta_n}$. We calculate the preference margin for the $i$th sample by 
\begin{align}
\mathbf p_m^{(i)} &= R_{\theta_m}(x_w^{(i)}) - R_{\theta_m}(x_l^{(i)}), \\
\mathbf p_n^{(i)} &= R_{\theta_n}(x_w^{(i)}) - R_{\theta_n}(x_l^{(i)}).
\end{align}
We group the preference margins of different samples as vectors, denoted by $\mathbf p_m$ and $\mathbf p_n$. HSIC assumes that the entries in $\mathbf p_m$ and $\mathbf p_n$ come from their respective underlying distributions, and the goal is to derive an empirical estimator for the dependence between the two distributions.

\textbf{HSIC dependency measure.} 
While the notion of covariance is a standard measure of dependency, it is restricted to linear correlation. The HSIC provides a more general way of capturing non-linear dependency by analyzing the relationship between variables in a high-dimensional feature space through kernel methods.

The HSIC begins by projecting the random variables $p_m$ and $p_n$ from their input space $\mathcal{X}$ onto a Reproducing Kernel Hilbert Space (RKHS), denoted $\mathcal{F}$, via a feature map $\phi(\cdot):\mathcal{X}\to\mathcal{F}.$ For an inner-product of $\phi(a)$ and $\phi(b)$, a computationally tractable kernel function is used
\begin{align}
    \langle \phi(a),\,\phi(b)\rangle_{\mathcal{F}}=k(a,b).
\end{align}
Thus, we do not need to explicitly compute $
\phi(\cdot)$. A common choice for the kernel is the Gaussian RBF kernel, $k(a,b)=\exp\!\left(-\frac{\|a-b\|^2}{2\sigma^2}\right)$, where the hyperparameter $\sigma$ controls its sensitivity. 

Within this framework, the concept of covariance is generalized by the cross-covariance operator:
\begin{align}
    C_{mn} \;=\; \mathbb{E}_{p_m,p_n}\!\left[ \big(\phi(p_m)-\mu_m\big)\otimes\big(\phi(p_n)-\mu_n\big) \right],
\end{align}
where $\mu_m=\mathbb{E}_{p_m}\!\big[\phi(p_m)\big]$ and $\mu_n=\mathbb{E}_{p_n}\!\big[\phi(p_n)\big]$ are the means. The tensor product $\otimes$ captures the complete covariance structure between the feature vectors.

HSIC is then defined as the squared Hilbert--Schmidt norm of this operator, which quantifies its magnitude:
\begin{align}
    \text{HSIC}(p_m, p_n) &= \|C_{mn}\|^2_{HS} \\
    &= \left\| \mathbb{E}_{p_m,p_n}\!\left[ \big(\phi(p_m)-\mu_m\big)\otimes\big(\phi(p_n)-\mu_n\big) \right] \right\|^2_{HS}
\end{align}
While the theoretical definition of HSIC is based on expectations over a distribution, in practice
we must compute an empirical estimate from a finite set of data samples. Let $\{(p_m^{(i)},\,p_n^{(i)})\}_{i=1}^b$ be $b$-many pairs of samples in a batch.
The kernel matrices for the two variables are
$\mathbf K_m \in \mathbb{R}^{b\times b}$ and $\mathbf K_n \in \mathbb{R}^{b\times b}$, given by
\begin{align}
 (\mathbf K_m)_{ij} = k\!\big(p_m^{(i)},\,p_m^{(j)}\big), 
 \quad\quad
 (\mathbf K_n)_{ij} = k\!\big(p_n^{(i)},\,p_n^{(j)}\big).
\end{align}

The empirical estimate for HSIC from this finite batch is given by a matrix formula:
\begin{align}
\operatorname{HSIC}_b(\mathbf p_m, \mathbf p_n)
  &= \frac{1}{(b-1)^2}\,\operatorname{tr}\!\big(\mathbf K_m\,\mathbf H\,\mathbf K_n\,\mathbf H\big),
\end{align}
where $\mathbf H \in \mathbb{R}^{b\times b}$ is the centering matrix, defined as
\begin{align}
\mathbf H \;=\; \mathbf I - \frac{1}{b}\,\mathbf{1}\mathbf{1}^{\mathsf T}.
\end{align}
Here, $\mathbf I$ is the identity matrix and $\mathbf{1}$ is a column vector of ones. Overall, the HSIC serves as a regularizer in the reward model training objective in Eqn.~(\ref{eq:joint_loss_margins}), offering a computationally feasible way to capture non-linear dependency, and thus promotes diversity in our learned reward models.

\section{Implementation Details\label{appendix:implement_details}}

\input{implement_details}

\paragraph{Reward model architecture.}
Base reward models use the \emph{AutoModelForSequenceClassification} architecture from \citet{wolf2020transformers}. The trainable parameters are a reward head and LoRA \citep{hu2022lora} adapters. The reward head is a two-layer feedforward network with $1024$ dimensions and a ReLU activation function, which follows the setup in~\citet {yang2024regularizing}. LoRA hyperparameters are listed in Tab.~\ref{tab:exp_details}. All reward model learning approaches employ the same architectures and have the same trainable parameters.

\textbf{Details of competing methods.} We discuss the competing methods as follows.
\begin{itemize}
\item\textbf{Baseline.} This is a vanilla model trained to minimize the objective in Eqn.~(\ref{eq:reward_obj}).

\item  \textbf{Training with labeled margin} \citep{wang2024secrets}.
This approach modifies the classic objective by enforcing that $y_w$ exceeds $y_l$ by a certain margin $m$:
\begin{equation}\label{eq:reward_loss_margin}
\mathcal{L}_{\text{margin}}(\theta)
= - \mathbb{E}_{(x, y_w, y_l) \sim D}\!\left[\log \sigma\!\left( r_\theta(x, y_w) - r_\theta(x, y_l) - m \right)\right].
\end{equation}
The Unified-Feedback dataset provides a margin label $m$ for each pairwise sample, which indicates the desired reward difference. Note that such an approach requires additional labels (i.e., a margin label for every sample) compared with our method, and thus it may not be applicable to other preference datasets. 

\item \textbf{Label smoothing} \citep{wang2024secrets}.
To account for potential annotation noise, this approach assumes that a labeled pair is flipped with probability of $\epsilon$ (set to $0.1$). Then the training objective thus becomes
\begin{align}\label{eq:reward_loss_labelsmooth}
\resizebox{\linewidth}{!}{$
\begin{aligned}
\mathcal{L}_{\text{smooth}}(\theta)
= - \mathbb{E}_{(x, y_w, y_l) \sim D}\!\left[
(1-\epsilon)\,\log \sigma\!\left( r_\theta(x, y_w) - r_\theta(x, y_l) \right)
+ \epsilon\,\log \sigma\!\left( r_\theta(x, y_l) - r_\theta(x, y_w) \right)
\right],
\end{aligned}
$}
\end{align}
This improves robustness to label errors and mitigates overfitting.

\item \textbf{Avg Ensemble} \citep{coste2024reward}. For this competing approach, 
we train three reward models (in the same way as the Baseline approach) with different random seeds and aggregate their outputs by simple averaging. 

\item  \textbf{GRM}~\citep{yang2024regularizing}. 
GRM regularizes reward model learning with a text generation objective:
\begin{align}\label{eq:sft}
\resizebox{1\linewidth}{!}{$
\begin{aligned}
\mathcal{L}_{\text{GRM}}(\theta, \theta_{\text{LM}}) = -\mathbb{E}_{(x, y_w, y_l) \sim D} \left[\log \left( \sigma \left( r_\theta(x, y_w) - r_\theta(x, y_l) \right) \right) \right] - 
\alpha \mathbb{E}_{(x, y_c) \sim D} \left[ \log \sigma \left( \beta \log \left( {\pi_{\theta_{\rm LM}}(y_c \mid x)} \right)  \right) \right].
\end{aligned}
$}
\end{align}

where we set $\beta = 0.1$ and $\alpha = 0.0005$ following the best setting reported by \citet{yang2024regularizing}.

\end{itemize}

\paragraph{Computing resources.}
Experiments are run on NVIDIA RTX~A6000 (48\,GB) GPUs. Training a 2B-parameter reward model with LoRA \citep{hu2022lora} on 40K training examples for two epochs requires approximately 31.9 GPU--hours.

\section{Additional Results}\label{sec:efficiency}

\subsection{Results with a Different Data Size}
\input{400k_main}
Following the experimental setup of \citet{yang2024regularizing}, we trained our reward models on the Unified-Feedback dataset with both 40K and 400K samples, as described in \S\ref{sec:reward_model_performance}. In the main paper, we only present the 40K-sample experiment (Tab.~\ref{tab:reward_performance_40k}) due to the limit of space. In this appendix, we show the results of the 400K-sample experiment in Tab.~\ref{tab:reward_performance_400k}. We see that our \methodname{} model surpasses all competing methods on both in-distribution and out-of-distribution (OOD) datasets. These findings are consistent with the performance observed in the 40K-sample experiment (Tab.~\ref{tab:reward_performance_40k}).

\subsection{Results with Full Training Data and a Larger Backbone}
\label{app:full_data_scaling}

\begin{table}[t]
    \centering
    \begin{minipage}[t]{0.49\textwidth}
        \centering
        \small
        \resizebox{\linewidth}{!}{
        \begin{tabular}{lcccc}
            \toprule
            Approach & Unified-Feedback & HHH-Align & MT-Bench & RewardBench \\
            \midrule
            Baseline RM              & 71.9 & 74.3 & 72.8 & 68.3 \\
            Label Smoothing          & 72.1 & 71.2 & 71.4 & 72.2 \\
            Margin Loss              & 71.9 & 76.5 & 73.2 & 72.6 \\
            GRM                      & 73.5 & 79.1 & 74.7 & 70.4 \\
            Avg Ensemble ($N=3$)     & 73.6 & 77.7 & 73.4 & 71.8 \\
            \textbf{RMB ($N=3$)}     & \textbf{75.8} & \textbf{80.5} & \textbf{75.9} & \textbf{73.0} \\
            \bottomrule
        \end{tabular}
        }
        \captionof{table}{Preference prediction performance of all approaches trained on the full Unified-Feedback dataset using the Gemma-2B-IT backbone.}
        \label{tab:full_data_gemma}
    \end{minipage}
    \hfill
    \begin{minipage}[t]{0.49\textwidth}
        \centering
        \small
        \resizebox{\linewidth}{!}{
        \begin{tabular}{lcccc}
            \toprule
            Approach & Unified-Feedback & HHH-Align & MT-Bench & RewardBench \\
            \midrule
            Baseline RM              & 76.1 & 87.5 & 76.3 & 76.4 \\
            Label Smoothing          & 76.7 & 88.0 & \textbf{78.1} & 77.3 \\
            Margin Loss              & 75.3 & 86.8 & 76.0 & 77.2 \\
            GRM                      & 76.9 & 87.6 & 77.0 & 78.6 \\
            Avg Ensemble ($N=3$)     & 76.4 & 88.3 & 76.8 & 78.9 \\
            \textbf{RMB ($N=3$)}     & \textbf{78.2} & \textbf{89.4} & 77.5 & \textbf{81.6} \\
            \bottomrule
        \end{tabular}
        }
        \captionof{table}{Preference prediction performance of all approaches trained on the full Unified-Feedback dataset using the Mistral-7B backbone.}
        \label{tab:full_data_mistral}
    \end{minipage}
\end{table}
\revision{
We further evaluate RMB with a larger training set and a larger reward-model backbone. Specifically, we train all approaches from scratch on the full Unified-Feedback training set, which contains 848,574 preference examples. We first use Gemma-2B-IT as the backbone, following the setup of our main experiments. We then scale the reward-model backbone to Mistral-7B and repeat the same experiment. For both settings, we evaluate preference prediction performance on Unified-Feedback, HHH-Alignment, MT-Bench, and RewardBench.

As shown in Tabs.~\ref{tab:full_data_gemma} and~\ref{tab:full_data_mistral}, RMB maintains strong performance when trained with the full Unified-Feedback dataset. With Gemma-2B-IT, RMB achieves the best performance across all four evaluation datasets. With Mistral-7B, RMB achieves the best performance on three of the four datasets and remains competitive on MT-Bench. These results show that the effectiveness of RMB holds with both substantially more training data and a larger reward-model backbone.

\paragraph{Computing resources.}
Training all approaches on the full Unified-Feedback dataset requires approximately 3,562 NVIDIA L40S GPU-hours with the Gemma-2B-IT backbone and 11,397 NVIDIA L40S GPU-hours with the Mistral-7B backbone.
}
\subsection{Results of Avoiding LLM-as-Judge's Inherent Biases}
\input{win_rate_gpt4o}

LLM may have its own stylistic preferences (e.g., verbosity, tone, formatting), which could introduce systematic bias in LLM-as-judge evaluation. To reduce this risk, we intentionally choose a judge from a different model family than the policy backbone (DeepSeek-R1 as judge vs.\ Gemma as generator) as shown in \S\ref{sec:reward_model_performance} and Tab.~\ref{tab:llm_eval}. This mitigates potential judge--generator coupling where the judge may implicitly favor outputs that resemble its own generation style or training priors, and thus helps avoid inflating the win rate due to family-specific alignment rather than reward quality. Moreover, all candidate policies are evaluated by the same judge under the same prompts and identical pairwise protocol, so any fixed judge bias is applied consistently across methods, and the comparison remains controlled.

To further check that the conclusion is not specific to DeepSeek-R1, we additionally evaluate with ChatGPT-4o as the judge using the exact same win/tie/lose methodology. The results in Tab.~\ref{tab:llm_eval_4o} confirm the same trend as Tab.~\ref{tab:llm_eval}, showing RMB consistently outperforms the baselines in policy training using different LLMs for evaluation.

\subsection{Analysis of Robustness to Prompt-Dependent Reward Shifts}
\label{app:shift_robustness}

\revision{
Our RM training uses 
the Bradley--Terry objective, which cannot identify the absolute reward values; i.e., the reward function may be shifted by a prompt-dependent constant. As mentioned in Section~\ref{sec:limitation}, our RMB does not guarantee preservation of preference order under such shifts. However, this is not a major threat to our method, because we train RMs first, and then train the boosting model given the RMs being fixed. Thus, the boosting model can adapt accordingly.

Nevertheless, we examine the sensitivity of RMB when such a constant is deliberately introduced at the inference stage after training the boosting model.

\paragraph{Mimicking prompt-dependent constants.}
We instantiate the prompt-dependent constants by independently sampling from a Gaussian distribution. To make the perturbation scale comparable with the raw scores of each RM, we consider the standard deviation of our trained RMs. Specifically, for RM $i$, we compute $\sigma_i$ over all preferred and disfavored responses in the test set. Then, the prompt-dependent constants are sampled
\begin{align}
c_i(x) \sim \mathcal{N}(0,\sigma_i^2), \qquad
\mathbf{C}(x)=[c_1(x),c_2(x),c_3(x)].
\label{eq:sample_prompt_shift}
\end{align}
Then, we apply the same $\mathbf{C}(x)$ to the preferred and disfavored responses:
\begin{align}
\mathbf{R'}(x,y_w) &= \mathbf R(x,y_w)+\mathbf{C}(x), \qquad
\mathbf R'(x,y_l) = \mathbf R(x,y_l)+\mathbf{C}(x).
\label{eq:prompt_shift}
\end{align}
With such a shift, we test the consistency of preference order of our RMB by using $\mathbf R$ or $\mathbf R'$.

\paragraph{Experiments.}
We conduct the experiment on the \texttt{llm-blender/Unified-Feedback} test set containing 7,289 preference pairs. In the experiment, the XGBoost aggregator is trained on the original RMs and is fixed when we add the sampled constants to RMs.
We show the statistics of reward scores and the sampled constants in Tab.~\ref{tab:prompt_shift_statistics}. 

\begin{table}[t]
\centering
\small
\begin{tabular}{lrrrrr}
\toprule
RM & Reward Mean & Reward Std. & $c(x)$ Mean  & $c(x)$ Std. & $c(x)$ P05--P95 \\
\midrule
RM 0 & 2.6380  & 17.0647 & -0.1391 & 17.0468 & [-28.2891, 27.9354] \\
RM 1 & 0.6850  & 10.9031 &  0.0566 & 10.9953 & [-17.9596, 18.3358] \\
RM 2 & -0.0619 & 10.9822 &  0.0606 & 10.9897 & [-17.8373, 17.7582] \\
\bottomrule
\end{tabular}
\caption{Statistics of reward scores and sampled prompt-dependent constants.}\bigskip

\label{tab:prompt_shift_statistics}

\centering
\small
\begin{tabular}{lrrrr}
\toprule
Dataset & Test Pairs & Rank Flips & Flip Ratio & \\
\midrule
Unified-Feedback & 7,289 & 297 & 4.07\%  \\
\bottomrule
\end{tabular}
\caption{Sensitivity of RMB rankings to prompt-dependent additive shifts.}
\label{tab:prompt_shift_rankflip}
\end{table}

We show the number and ratio of flipped pairwise order in Tab.~\ref{tab:prompt_shift_rankflip}. As seen, only 297 out of 7,289 preference orders are changed in the RMB ranking, corresponding to a rank flip ratio of $4.07\%$. This indicates that RMB's rankings remain largely stable if RMs are shifted.
}

\section{Training Efficiency}
\label{app:training-efficiency}
This appendix presents a comprehensive analysis of the computational efficiency of \methodname, detailing the breakdown of training costs and examining the trade-off between performance and efficiency as the number of reward models, $N$, increases. We demonstrate that \methodname{} incurs a training cost comparable to existing ensemble-based approaches and that the method remains robust across reasonable choices of $N$.

\paragraph{Reward-Model Training Cost}
The primary computational overhead in \methodname{} stems from training the base RMs. As the number of RMs increases, the training cost grows linearly. In our experimental setup, training a single RM per epoch of data requires approximately 31.9 GPU-hours on an NVIDIA RTX A6000 (48GB). However, the most appropriate comparison is not against a single, weak RM, but rather against other methods that employ multiple RMs to mitigate reward hacking, such as reward model ensembles \citep{coste2024reward, eisenstein2024helping}. When compared under a fixed budget of $N$ trained RMs, \methodname{} proves to be both efficient and effective. For example, with $N=3$, \methodname{} outperforms an average ensemble using $N=3$ or even $N=5$ on both preference prediction and downstream RLHF tasks. This indicates that \methodname{} achieves superior performance with lower total computational expenditure than competing ensemble-based methods, which often require training a larger number of models to achieve similar or weaker gains.

\paragraph{Cost of HSIC Diversity Regularization}
\methodname{} jointly trains $N$ RMs with a diversity-promoting regularizer based on the HSIC. This term is computed over per-batch margin vectors and is implemented via a small number of matrix operations. In practice, the computational overhead of this regularization is negligible relative to the forward and backward passes required for the $N$ large language models serving as RMs.

\paragraph{Boosting Aggregator Training Cost}
Following the training of the RMs, \methodname{} learns a tree-based boosting aggregator on the frozen RM outputs. This stage is extremely lightweight; in our experiments, training the aggregator on the full dataset requires less than 5 minutes on a CPU. This is negligible compared to the GPU-hours required for RM training. Consequently, the total training cost of \methodname{} remains on the same order as existing ensemble methods while delivering substantially higher accuracy and robustness.

\paragraph{Performance--Training Efficiency Trade-Off}

\input{tradeoff_plot}

Fig.~\ref{fig:tradeoff} illustrates the preference prediction accuracy against the total training cost (in GPU-hours per GPU) for varying ensemble sizes $N \in \{1,3,5,8,10\}$. Several key trends emerge from this analysis. First, we observe monotonic improvements for small $N$; increasing $N$ from $1$ to $5$ yields consistent accuracy gains across both in-distribution and out-of-distribution evaluation sets, with training costs growing roughly linearly. Second, performance gains begin to saturate and exhibit mild overfitting as $N$ increases from $8$ to $10$. This plateau suggests that as the input dimension (the vector of $N$ RM scores) grows, an overly flexible aggregator may begin to fit spurious correlations or idiosyncratic noise within the specific collection of RMs rather than learning a generalizable aggregation rule, which suggests overfitting due to an over-powerful aggregator model. Finally, the performance curve reveals a wide and relatively flat plateau for $N \in [5,10]$. This stability indicates that \methodname{} is not sensitive to precise tuning of $N$, as a broad range of values yields similarly strong results without sharp performance drops.

\section{Choosing the Number of Reward Models $N$}
\label{app:choose-N}

This appendix provides a principled perspective on selecting the ensemble size $N$ in \methodname{}.
Increasing $N$ can improve robustness by aggregating information across multiple reward models, but the gains diminish when additional models contribute little \emph{independent} signal.
Moreover, a larger ensemble increases the input dimensionality of the downstream boosting aggregator, which can raise the risk of overfitting.
We formalize these effects using a standard correlated-error model, derive an \emph{effective ensemble size} that makes saturation explicit.

\subsection{A Correlated-Error Model for Ensemble Margins}
\label{app:choose-N:model}

To quantify diminishing returns with increasing $N$, we adopt a classical exchangeable (equicorrelated) error model \citet{krogh1994neural, donner2000design,buhlmann2002analyzing,kuncheva2014combining}.
Write each margin predictor in Eqn.~(\ref{eq:margin_vec}) as
\begin{equation}
m_k(x) \;=\; m^\star(x) + \varepsilon_k(x),
\label{eq:signal-plus-noise-app}
\end{equation}
where $m^\star(x)$ is a latent margin signal and $\varepsilon_k(x)$ is the model-specific error.
For tractability, we assume that for any fixed $x$:
\begin{align}
\mathbb{E}[\varepsilon_k(x)] &= 0, \\
\mathrm{Var}(\varepsilon_k(x)) &= \sigma^2 \qquad \text{for all } k, \\
\mathrm{Corr}(\varepsilon_i(x), \varepsilon_j(x)) &= \rho \qquad \text{for all } i\neq j.
\label{eq:equicorr-assumption-app}
\end{align}

\subsection{Variance of the Averaged Margin and Diminishing Returns}
\label{app:choose-N:variance}

Define the averaged margin vector over N RMs:
\begin{equation}
\bar m_N(x) \;\triangleq\; \frac{1}{N}\sum_{k=1}^N m_k(x)
\;=\; m^\star(x) + \bar\varepsilon_N(x),
\qquad
\bar\varepsilon_N(x) \triangleq \frac{1}{N}\sum_{k=1}^N \varepsilon_k(x).
\label{eq:avg-margin-app}
\end{equation}
Since $m^\star(x)$ is fixed given $x$, the variance of $\bar m_N(x)$ is the variance of $\bar\varepsilon_N(x)$.
We compute $\mathrm{Var}(\bar\varepsilon_N)$ explicitly:
\begin{align}
\mathrm{Var}\!\left(\bar\varepsilon_N\right)
&= \mathrm{Var}\!\left(\frac{1}{N}\sum_{k=1}^N \varepsilon_k\right)
= \frac{1}{N^2}\mathrm{Var}\!\left(\sum_{k=1}^N \varepsilon_k\right) \nonumber\\
&= \frac{1}{N^2}\left(\sum_{k=1}^N \mathrm{Var}(\varepsilon_k) \;+\; 2\sum_{1\le i<j\le N}\mathrm{Cov}(\varepsilon_i,\varepsilon_j)\right) \nonumber\\
&= \frac{1}{N^2}\left(N\sigma^2 + 2\binom{N}{2}\rho\sigma^2\right)
= \sigma^2\frac{1+(N-1)\rho}{N}.
\label{eq:var-avg-app}
\end{align}
An equivalent and often more interpretable form is
\begin{equation}
\mathrm{Var}\!\left(\bar\varepsilon_N\right)
= \sigma^2\left(\rho + \frac{1-\rho}{N}\right).
\label{eq:var-avg-floor-app}
\end{equation}
Eqn.~(\ref{eq:var-avg-floor-app})
highlights two regimes.
If $\rho\approx 0$ (nearly independent errors), the variance decays as $\sigma^2/N$.
If $\rho>0$, the variance approaches the non-vanishing floor $\sigma^2\rho$ as $N\to\infty$, and gains from increasing $N$ must eventually saturate.

\subsection{Effective Ensemble Size $N_{\mathrm{eff}}$}
\label{app:choose-N:Neff}

A useful summary of the correlation penalty is the \emph{effective ensemble size} $N_{\mathrm{eff}}$, defined as the size of an \emph{independent} ensemble ($\rho=0$) that would achieve the same variance as the correlated ensemble.
For independent errors, $\mathrm{Var}(\bar\varepsilon)=\sigma^2/N_{\mathrm{eff}}$.
Equating this to Eqn.~(\ref{eq:var-avg-app}) yields
\begin{equation}
\frac{\sigma^2}{N_{\mathrm{eff}}}
= \sigma^2\frac{1+(N-1)\rho}{N}
\quad \Longrightarrow \quad
N_{\mathrm{eff}} \;=\; \frac{N}{1+(N-1)\rho}.
\label{eq:Neff-app}
\end{equation}
This expression makes the dependence-limited regime explicit:
when $\rho>0$, $N_{\mathrm{eff}}$ grows sublinearly and satisfies $N_{\mathrm{eff}}\to 1/\rho$ as $N\to\infty$.
In other words, adding more reward models cannot yield unbounded benefits unless their errors are close to independent.

\subsection{A Guideline for Selecting $N$}
\label{app:choose-N:guideline}

Eqn.~(\ref{eq:var-avg-floor-app}) shows that the reducible portion of the averaged-error variance is
\begin{equation}
\mathrm{Var}(\bar\varepsilon_N) - \sigma^2\rho
= \sigma^2\frac{1-\rho}{N}.
\label{eq:reducible-variance-app}
\end{equation}
This motivates selecting $N$ so that the remaining reducible term is below a user-chosen tolerance.
For a tolerance $\tau>0$,
\begin{equation}
\sigma^2\frac{1-\rho}{N} \le \tau
\quad \Longrightarrow \quad
N \ge \frac{\sigma^2(1-\rho)}{\tau}.
\label{eq:N-from-tolerance-app}
\end{equation}
Because $\sigma^2$ is a scale factor, a convenient alternative is a \emph{relative} tolerance $\eta\in(0,1)$ defined with respect to the reducible term at $N=1$:
\begin{equation}
\frac{\mathrm{Var}(\bar\varepsilon_N)-\sigma^2\rho}{\mathrm{Var}(\bar\varepsilon_1)-\sigma^2\rho}
= \frac{1}{N}
\le \eta
\quad \Longrightarrow \quad
N \ge \frac{1}{\eta}.
\label{eq:N-relative-tolerance-app}
\end{equation}
This makes the diminishing-returns phenomenon explicit: once $N$ is moderately large, further increases can only shrink the remaining reducible term by a factor proportional to $1/N$, while the correlation floor $\sigma^2\rho$ is unchanged.

\subsection{Estimating Dependence on Held-out Data}
\label{app:choose-N:rho-est}

In practice, $\rho$ is unknown but can be approximated using held-out margins.
Let $\mathcal{D}_{\mathrm{val}}$ be a validation set of preference pairs.
For each model $k$, form the margin vector
\begin{equation}
\mathbf{m}_k \;=\; \big(m_k(x)\big)_{x\in \mathcal{D}_{\mathrm{val}}}\in\mathbb{R}^{|\mathcal{D}_{\mathrm{val}}|}.
\label{eq:margin-vector-app}
\end{equation}
A simple proxy for inter-model dependence is the average pairwise correlation of centered margin vectors:
\begin{equation}
\hat\rho \;\triangleq\; \frac{2}{N(N-1)}\sum_{i<j}
\mathrm{Corr}\!\left(\mathbf{m}_i-\bar{\mathbf{m}},\ \mathbf{m}_j-\bar{\mathbf{m}}\right),
\qquad
\bar{\mathbf{m}} \triangleq \frac{1}{N}\sum_{k=1}^N \mathbf{m}_k.
\label{eq:rho-hat-app}
\end{equation}
This estimate focuses on shared variation across examples rather than shared offsets.
Using $\hat\rho$, one can compute $\widehat{N}_{\mathrm{eff}}(N)=\frac{N}{1+(N-1)\hat\rho}$ as a diagnostic: a plateau in $\widehat{N}_{\mathrm{eff}}$ indicates that additional reward models are largely redundant.

\subsection{Interaction with the Boosting Aggregator and a Practical Selection Procedure}
\label{app:choose-N:procedure}

The analysis above quantifies variance reduction under averaging.
RMB uses a learned boosting aggregator rather than a simple mean, which makes validation-based selection particularly important.
As $N$ increases, the aggregator receives more features and may fit spurious interactions among partially redundant reward-model outputs, even with regularization.
Therefore, we recommend selecting $N$ using a two-criterion procedure that mirrors the theory:

\paragraph{(1) Diversity criterion (dependence-limited).}
Increase $N$ only while $\widehat{N}_{\mathrm{eff}}(N)$ continues to grow meaningfully. Stop when
\begin{equation}
\frac{\widehat{N}_{\mathrm{eff}}(N+1)-\widehat{N}_{\mathrm{eff}}(N)}{\widehat{N}_{\mathrm{eff}}(N)} < \delta,
\label{eq:Neff-stop-app}
\end{equation}
for a small $\delta\in[0.01,0.05]$.

\paragraph{(2) Generalization criterion (complexity-limited).}
For each candidate $N$, train the boosting aggregator with early stopping on $\mathcal{D}_{\mathrm{val}}$ using the same pairwise objective as in RMB.
Increase $N$ only if the \emph{validation} objective improves by at least $\epsilon$ (or if the improvement is statistically reliable under bootstrap resampling over validation examples).

This procedure is computationally efficient in RMB because reward-model training typically dominates cost and scales approximately linearly with $N$.
In practice, one can train reward models incrementally (e.g., add 1--2 models at a time), cache their margins on $\mathcal{D}_{\mathrm{val}}$, and re-fit only the lightweight boosting aggregator during selection.

\section{HSIC Ratio Analysis}

The selection of the diversity regularization coefficient, $\lambda_{HSIC}$, follows the established practice of balancing the main task loss with the diversity penalty ~\citep{dong2023diversity, Ma2024TowardsCL}. In this appendix, we perform a sensitivity analysis by sweeping $\lambda_{HSIC}$ across four orders of magnitude (from $0.001$ to $10$) to evaluate its impact on model performance. We report results on the Unified-Feedback dataset, utilizing the same experimental setup detailed in Section~\ref{sec:indepth_analysis}.

\input{sensitivity}
As presented in Tab.~\ref{tab:sensitivity}, our approach demonstrates strong robustness to the choice of $\lambda_{HSIC}$. The performance of RMB remains stable and effective across a broad range of values ($0.001 \le \lambda_{HSIC} \le 1$), even as the average pairwise correlation among reward models varies significantly (from $92\%$ down to $59\%$). This indicates that RMB can effectively leverage increasing diversity to boost performance, provided the base estimators maintain reasonable accuracy. Performance deterioration is observed only when the regularization becomes overwhelmingly dominant ($\lambda_{HSIC}=10$). In this regime, while diversity is maximized (correlation drops to $51\%$), the aggressive penalty interferes with the optimization of the main reward objective. Consequently, the accuracy of individual reward models degrades sharply ($61.7\%$), which in turn limits the effectiveness of the boosted aggregator.

\section{Diversity Regularizer Selection}
\label{app:diversity-regularizer}

In this appendix, we present an additional analysis for comparing the HSIC regularizer with two other widely used diversity-promoting objectives:  Negative Correlation Learning (NCL) ~\citep{liu1999ensemble}  and Determinantal Point Processes (DPP)~\citep{taskar2013determinantal}.

We prioritize HSIC as the diversity-promoting regularizer in the reward model learning scenario for two reasons. First, HSIC allows for direct control of statistical dependence by measuring and penalizing the dependence between the margin vectors of different RMs. This is precisely the form of diversity required for effective boosting-based aggregation. Second, the HSIC objective is bounded and scale-stable. The empirical HSIC values are strictly bounded between $0$ and $1$, ensuring a well-behaved scale when combined with the primary reward-learning objective. In contrast, losses based on DPP or NCL are theoretically unbounded from above, which complicates the selection of trade-off coefficients and can destabilize the optimization process.

We conducted a pilot study where we replaced the HSIC term with either DPP or NCL while keeping all other settings identical for $N=5$ RMs. Tab.~\ref{tab:diversity-ablations} reports the resulting preference prediction accuracies on the Unified-Feedback dataset.

\begin{table}[t]
\centering
\begin{tabular}{lccccc}
\toprule
Regularizer & RM1 & RM2 & RM3 & RM4 & RM5 \\
\midrule
HSIC & 73.40 & 70.73 & 69.29 & 68.05 & 73.84 \\
DPP  & 60.58 & 64.93 & 55.20 & 62.74 & 57.19 \\
NCL  & 56.02 & 59.45 & 52.87 & 57.40 & 55.16 \\
\bottomrule
\end{tabular}
\caption{Comparison of different diversity regularizers. We report the accuracy (\%) of five
individually trained reward models (RM1--RM5) under HSIC, DPP, and NCL regularization. HSIC
consistently yields stronger RMs, while DPP and NCL significantly underperform.}
\label{tab:diversity-ablations}
\end{table}

The results indicate that both DPP and NCL yield substantially lower accuracies across all five RMs. Furthermore, they proved more sensitive to hyperparameter tuning due to the unbounded magnitude of their loss terms. By contrast, HSIC facilitates the training of individually superior RMs while maintaining stable optimization dynamics. Based on these empirical findings, we adopt HSIC as the diversity regularizer for \methodname{} and exclude DPP and NCL from the main model design.

\section{LLM Evaluation Prompt Template}\label{apd:llm-eval} 
To evaluate the quality of the text generated by policies trained with different reward signals, we employ the LLM-as-judge paradigm, which has been widely adopted for automated evaluation of generative models \citep{zheng2023judging,liu2024provably,fan2026ketchup}. In particular, we prompt Deepseek-R1 \citep{deepseekai2025deepseekr1incentivizingreasoningcapability} to conduct both pairwise and pointwise assessments of the generated responses.
Our prompt templates originate from \citet{zheng2023judging}, and are shown in Tab.~\ref{tab:llm_prompt_pariwse} and \ref{tab:llm_prompt_pointwise}. To ensure the fairness and reliability of our pairwise comparisons, we adapt the prompt structure in \citet{fan2026ketchup}, which is designed to mitigate ID and position bias~\citep{zheng2023large,shen2023large}.

\begin{table}[!t]
\centering
\resizebox{.8\columnwidth}{!}{
\begin{tabular}{@{}l@{}}
\toprule
Please act as an impartial judge and evaluate the quality of the responses provided by two \\
AI assistants to the user question displayed below.
Your evaluation should consider \\
correctness and helpfulness. You will be given a reference answer, assistant A’s answer, \\
and assistant B’s answer. 
Your job is to evaluate which assistant’s answer is better. \\
Begin your evaluation by comparing both assistants’ answers with the reference answer.\\
Identify and correct any mistakes. Avoid any position biases and ensure that the order in \\
which the responses were presented does not influence your decision.
Do not allow the  \\
length of the responses to influence your evaluation. Do not favor certain names of the
\\ assistants. 
Be as objective as possible.
\\
\\
Source: \textbf{[Source]} \\
Answer \textbf{[ID1]}: \textbf{[Answer-A]} \\
Answer \textbf{[ID2]}:  \textbf{[Answer-B]} \\
\\

After providing your explanation, output your
final verdict by strictly following this format: \\
"[[ID1]]" if assistant A is better, "[[ID2]]"
if assistant B is better, and "[[ID-TIE]]" for a tie. \\
Your response should use the format: \\
\\
Overall Quality: one-sentence comparison and explanation \\
Preferred: [[Answer ID]] \\
\bottomrule
\end{tabular}}
\caption{Prompt templates for LLM evaluation on text generated by RLHF optimized policy in \S\ref{sec:reward_model_performance} from a pairwise view. This template originates from ~\cite{zheng2023judging}'s work, and is adapted according to~\cite{fan2026ketchup}'s template to avoid ID and position bias, aiming to provide reliable LLM judges for pairwise comparison.}
\label{tab:llm_prompt_pariwse}
\end{table}

\begin{table}[!t]
\centering
\resizebox{.8\textwidth}{!}{
\begin{tabular}{@{}l@{}}
\toprule
Please act as an impartial judge and evaluate the quality of the response provided by an
AI assistant \\
to the user’s question shown below. Your evaluation should consider factors
such as helpfulness, \\
relevance, accuracy, depth, creativity, and level of detail in
the response. \\
Begin your evaluation with a short explanation. Be as objective as
possible. \\
After providing your explanation, please rate the response on a scale of 1 to 10 \\

by strictly following this format: "[[rating]]", for example: "Rating: [[5]]".
\\[1ex]
\textbf{[Question]} \\
\{question\} \\
\textbf{[The Start of Assistant’s Answer]} \\
\{answer\} \\
\textbf{[The End of Assistant’s Answer]} \\
\bottomrule
\end{tabular}
}
\caption{Prompt templates for LLM evaluation on text generated by RLHF optimized policy in \S\ref{sec:reward_model_performance} from a pointwise view. This template originates from ~\cite{zheng2023judging}'s work, aiming to provide LLM judges for single-answer grading.}
\label{tab:llm_prompt_pointwise}
\end{table}

\section{Reproducibility Statement}
All code is provided via a GitHub repository, including implementations for data loading, reward model training, and policy optimization. The datasets used are publicly available, and we release the complete set of training hyperparameters. Our evaluation approaches are also publicly available and can be fully reproduced.

\end{document}

%% file: math_commands.tex
\usepackage{amsmath,amsfonts,bm}

\def\eqref#1{equation~\ref{#1}}

\def\1{\bm{1}}

\DeclareMathAlphabet{\mathsfit}{\encodingdefault}{\sfdefault}{m}{sl}
\SetMathAlphabet{\mathsfit}{bold}{\encodingdefault}{\sfdefault}{bx}{n}



%% file: 40k_main.tex
\definecolor{avgcol}{gray}{0.92}
\newcolumntype{A}{>{\columncolor{avgcol}}c} 

\begin{table*}[t]
\centering
\scriptsize
\setlength{\tabcolsep}{5pt}
\renewcommand{\arraystretch}{1.15}
\begin{tabular}{lcccAcccc}
\toprule
\multirow{2}{*}{\textbf{Approach}} &
\multirow{2}{*}{\textbf{Unified-Feedback}} &
\multirow{2}{*}{\textbf{HHH-Align}} &
\multirow{2}{*}{\textbf{MT-Bench}} &
\multicolumn{5}{c}{\textbf{RewardBench}} \\
\cmidrule(lr){5-9}
& & & & \multicolumn{1}{c}{Avg.} & Chat & Chat-Hard & Safety & Reasoning \\
\midrule
Frozen RM        & 63.9  & 68.6  & 68.2  & x     & x     & x     & x     & x     \\
Baseline RM      & 68.8  & 70.3  & 69.1  & 64.5  & 95.8  & 37.3  & 59.9  & 64.8  \\
Label Smooth  & 68.5  & 68.8  & 71.9  & 61.1  & 91.6  & 39.0  & 53.8  & 60.2  \\
Margin Loss  & 69.6  & 69.8  & 71.0  & 66.4  & {\bfseries 97.2}  & 37.5  & 56.8  & {\bfseries 72.7}  \\
GRM                  & 71.5  & 78.7  & 73.0  & 66.8  & 94.1  & 41.9  & 69.5  & 61.5  \\
Avg. Ensemble ($N=3$)      & 69.9  & 72.2  & 71.1  & 65.2  & 96.1  & 38.2  & 58.8  & 67.6  \\
Avg. Ensemble ($N=5$)      & 70.8  & 74.5  & 72.6  & 68.9  & 95.3  & 41.7  & 70.4  & 70.2  \\
\midrule
RMB ($N=3$)           & 73.6 &  81.3 & {\bfseries 76.3} & 68.2 &  96.4 & {\bfseries 44.1} &  72.4 & 66.0 \\
RMB ($N=5$)           & {\bfseries 74.6} & {\bfseries 83.2} & 76.0 &  {\bfseries 71.7} & 95.3 & 43.6 & {\bfseries 78.3} & 71.4 \\
\bottomrule
\end{tabular}
\caption{Preference prediction performance on all approaches learned using 40K data mode of Unified-Feedback.  Results of competing approaches are directly sourced from \citet{yang2024regularizing}.  $N$ means the number of reward models used for ensembling or boosting.}
\label{tab:reward_performance_40k}
\end{table*}

%% file: win_rate_pairwise.tex
\begin{table*}[t]
\vspace*{-0pt}
\centering
\resizebox{0.95\linewidth}{!}{
\begin{tabular}{c c c c c c c c c c}
\toprule
\multirow{2}{*}{\textbf{Approach}} &
\multirow{2}{*}{\textbf{Opponent}} &
\multicolumn{4}{c}{\textbf{Pairwise Preference Comparison} (in \%)} &
\multicolumn{4}{c}{\textbf{Pointwise Preference Comparison} (in \%)}\\
\cmidrule(lr){3-6}\cmidrule(lr){7-10}
& & \textbf{Win$\uparrow$} & \textbf{Tie} & \textbf{Lose$\downarrow$} & \boldmath{$\Delta$} &
\textbf{Win$\uparrow$} & \textbf{Tie} & \textbf{Lose$\downarrow$} & \boldmath{$\Delta$} \\
\midrule
\multirow{3}{*}{\textbf{RMB} ($N=3$)} 
 & Baseline RM   & 55.97 & 3.21 & 40.82 & $\uparrow$ \textbf{15.15} & 44.73 & 23.97 & 31.29 & $\uparrow$ \textbf{13.44} \\
 & Avg. Ensemble ($N=3$)  & 57.87 & 4.31 & 37.81 & $\uparrow$ \textbf{20.06} & 40.22 & 34.50 & 25.28 & $\uparrow$ \textbf{14.94} \\
 & GRM           & 54.76 & 3.01 & 42.22 & $\uparrow$ \textbf{12.54} & 43.33 & 30.39 & 26.28 & $\uparrow$ \textbf{7.05}  \\
\bottomrule
\end{tabular}
}
\caption{LLM-as-judge of policies learned via RLHF.}
\label{tab:llm_eval}
\end{table*}

%% file: bon.tex

\begin{figure*}[t]
  \centering
  \captionsetup[subfigure]{skip=2pt,margin=0pt,labelformat=parens}

  \begin{subfigure}[t]{0.235\textwidth}
    \includegraphics[width=\linewidth]{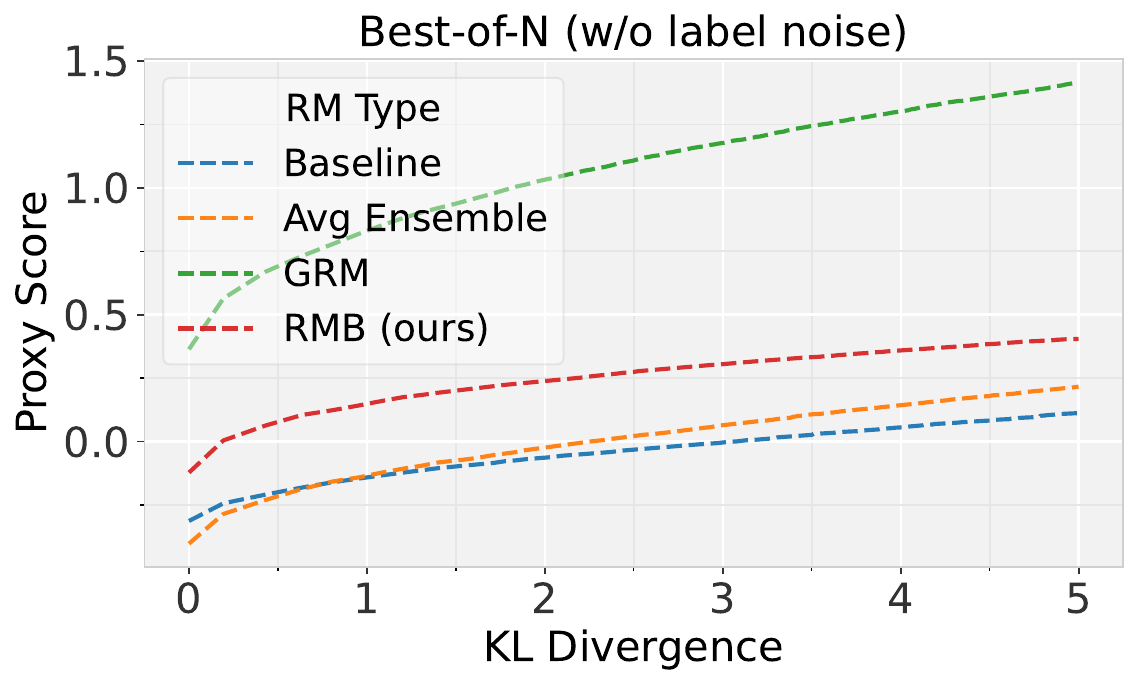} 
    \subcaption{}\label{fig:bon-a}
  \end{subfigure}
  \begin{subfigure}[t]{0.235\textwidth}
    \includegraphics[width=\linewidth]{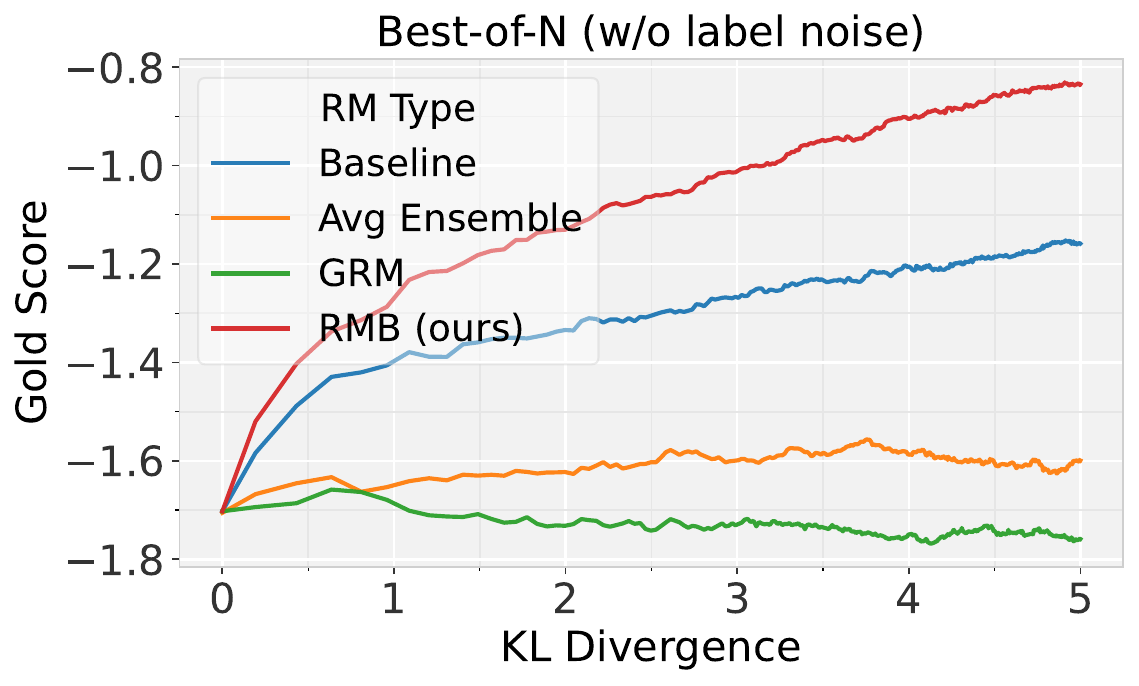}
    \subcaption{}\label{fig:bon-b}
  \end{subfigure}
  \begin{subfigure}[t]{0.235\textwidth}
    \includegraphics[width=\linewidth]{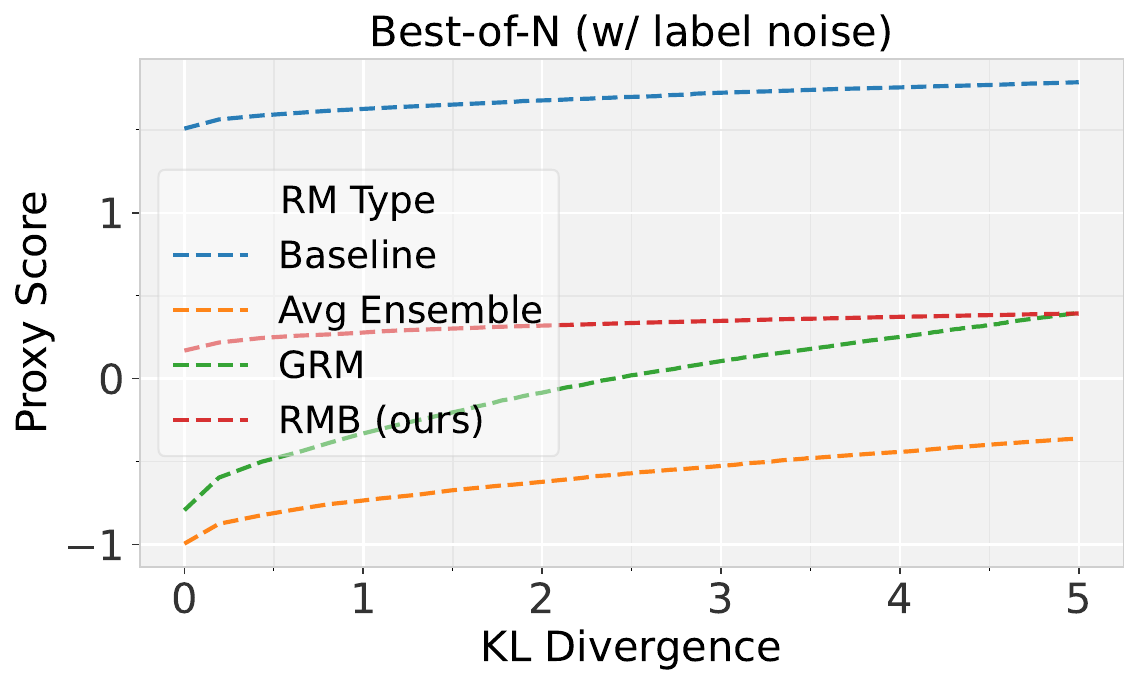}
    \subcaption{}\label{fig:bon-c}
  \end{subfigure}
  \begin{subfigure}[t]{0.235\textwidth}
    \includegraphics[width=\linewidth]{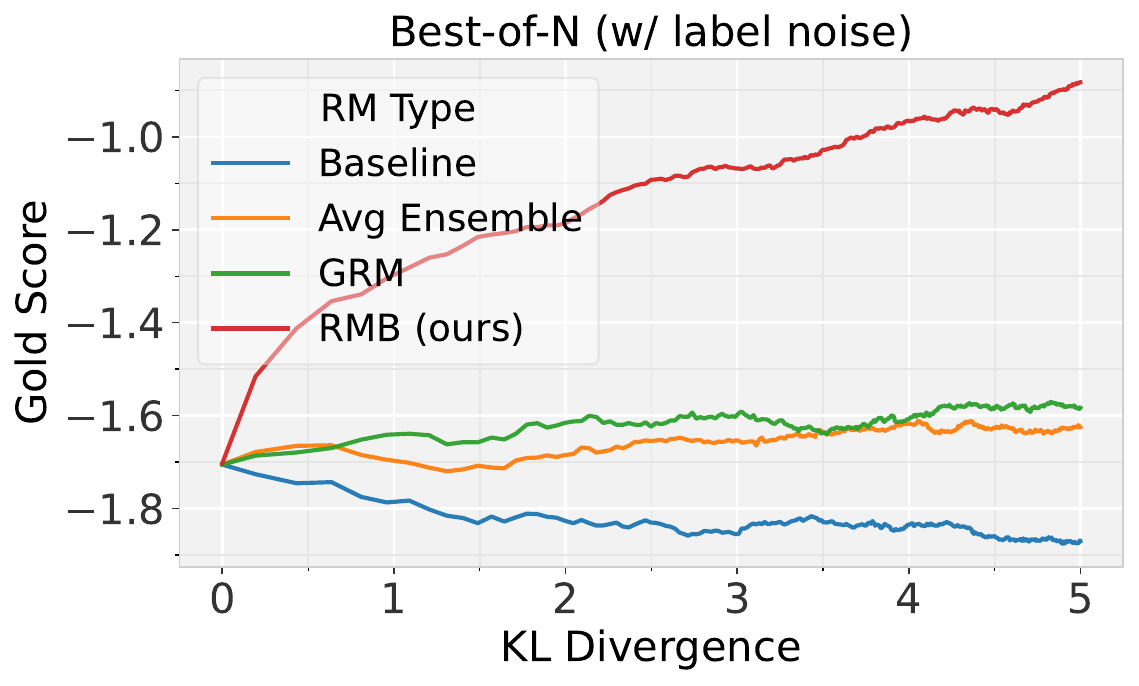}
    \subcaption{}\label{fig:bon-d}
  \end{subfigure}
  \caption{Results of Best-of-N sampling experiments, where both Avg Ensemble and our RMB use three individual reward models. Figures~\ref{fig:bon-a} and~\ref{fig:bon-c} intend to show the change of the proxy reward score, as an increasing proxy reward with a decreasing/plateaued gold reward indicates reward hacking. Notice that the scores of different proxy reward models are not directly comparable. Figures~\ref{fig:bon-b} and ~\ref{fig:bon-d} show the gold reward scores achieved, a higher score indicating a better reward model.}
  \label{fig:BoN}
\end{figure*}

%% file: ppo.tex

\begin{figure*}[t]
  \centering
  \captionsetup[subfigure]{skip=2pt,margin=0pt,labelformat=parens}

  \begin{subfigure}[t]{0.235\textwidth}
    \includegraphics[width=\linewidth]{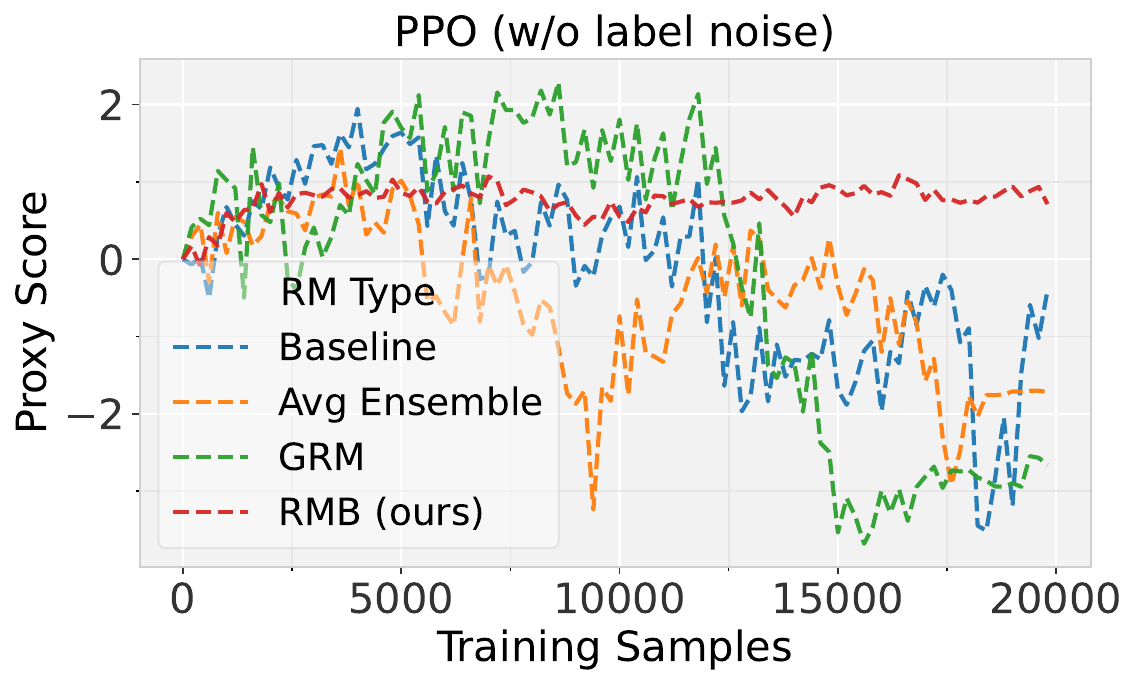} 
    \subcaption{}\label{fig:ppo-a}
  \end{subfigure}
  \begin{subfigure}[t]{0.235\textwidth}
    \includegraphics[width=\linewidth]{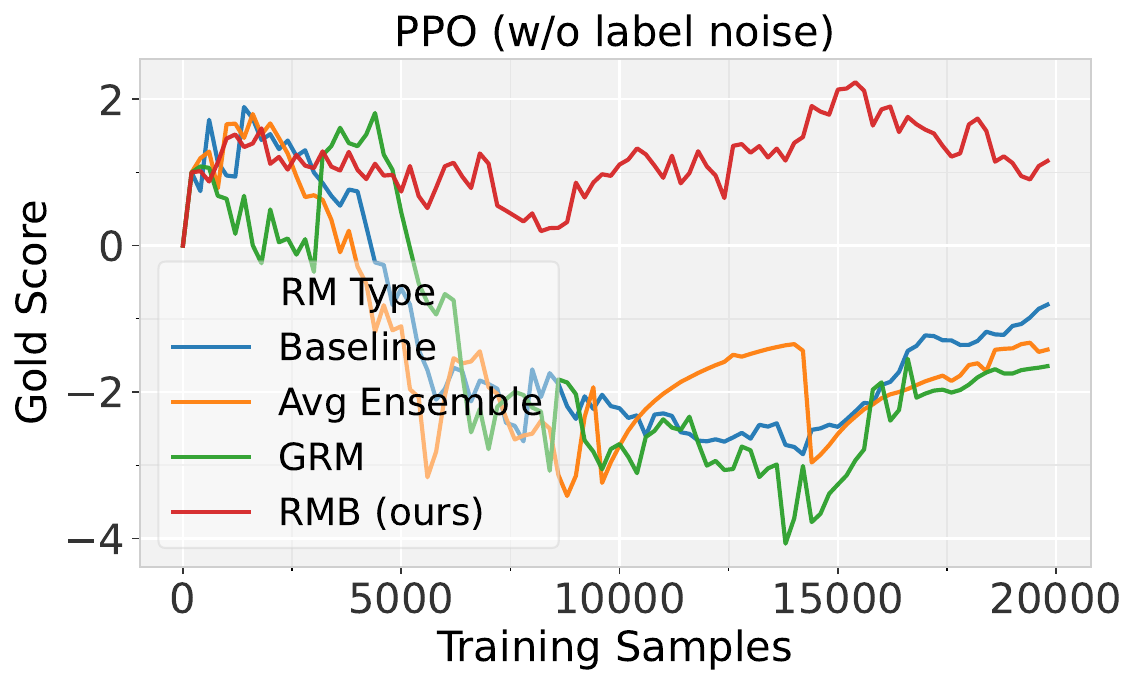}
    \subcaption{}\label{fig:ppo-b}
  \end{subfigure}
  \begin{subfigure}[t]{0.235\textwidth}
    \includegraphics[width=\linewidth]{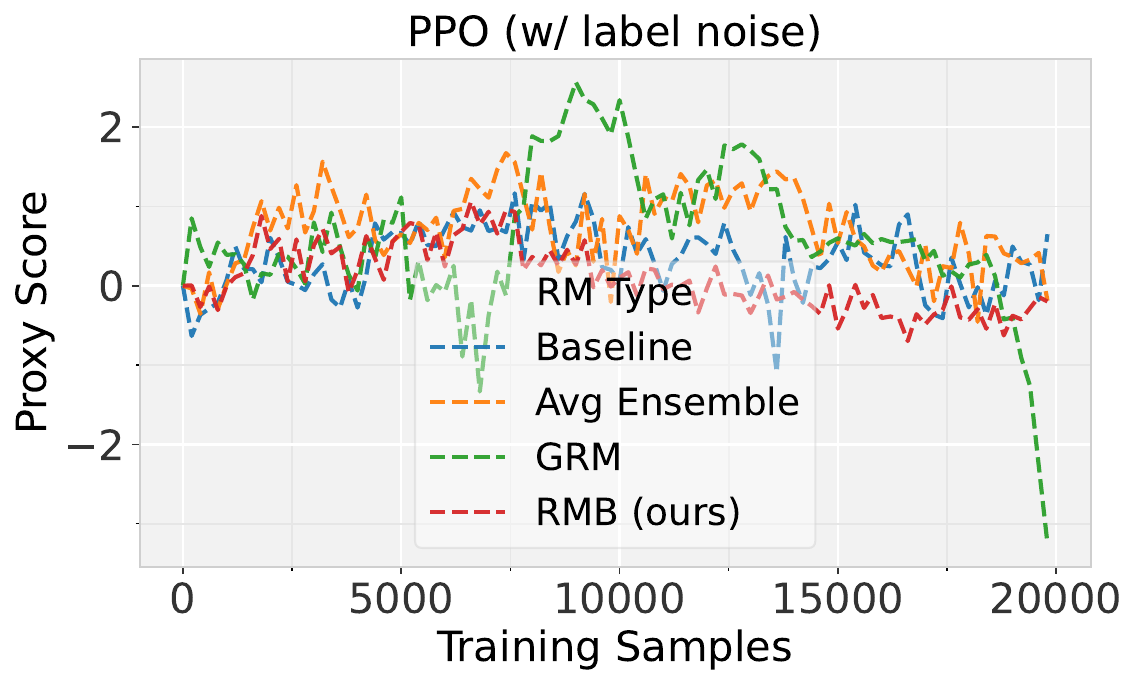}
    \subcaption{}\label{fig:ppo-c}
  \end{subfigure}
  \begin{subfigure}[t]{0.235\textwidth}
    \includegraphics[width=\linewidth]{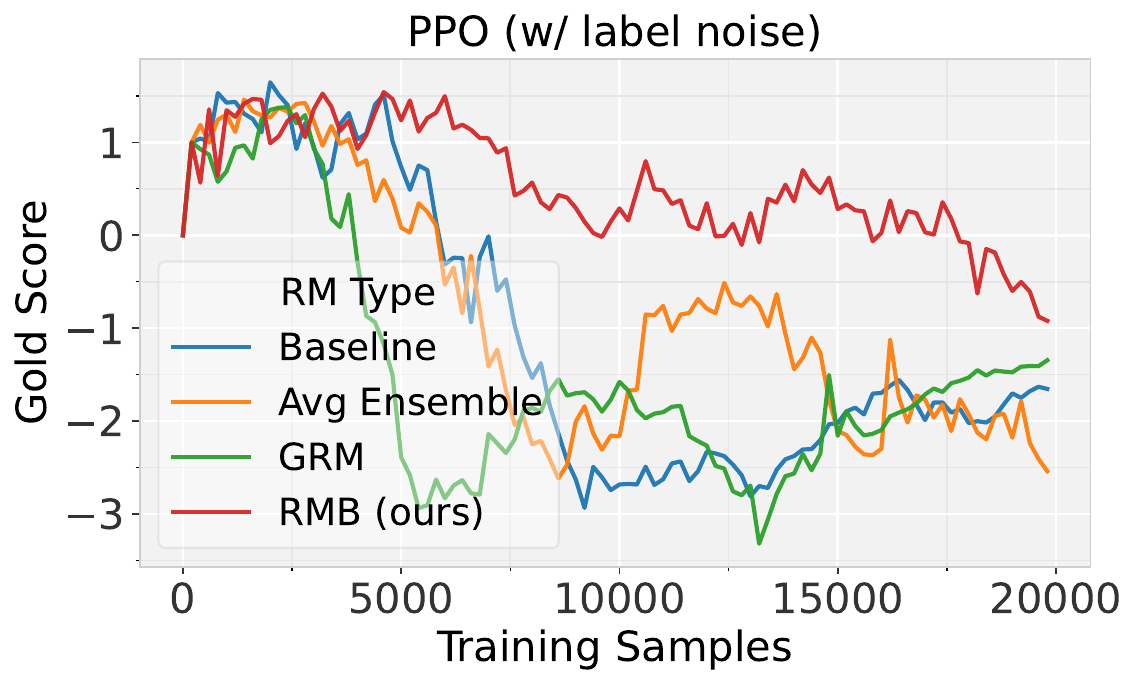}
    \subcaption{}\label{fig:ppo-d}
  \end{subfigure}
  \caption{PPO training with reward models (RMs). Ensemble and RMB both have three RMs to aggregate.}
  \label{fig:PPO}
\vspace*{-0pt}
\end{figure*}

%% file: ablation.tex
\begin{table*}[t]
\vspace*{-0pt}
\centering
\scriptsize
\setlength{\tabcolsep}{3.5pt}
\renewcommand{\arraystretch}{1.2}

\resizebox{0.9\linewidth}{!}{%
\begin{tabular}{|p{1.5cm}|l|*{5}{c|}*{5}{c|}}
\hline
\multicolumn{2}{|c|}{} &
\multicolumn{5}{c|}{\textbf{Unified-Feedback} (In-Distribution)} &
\multicolumn{5}{c|}{\textbf{RewardBench} (Out-of-Distribution)} \\
\hline
\multirow{3}{*}{\shortstack[c]{Random \\ Seeds}}
& Individual RMs      & 70.71 & 71.34 & 70.52 & 69.85 & 70.30 & 68.45 & 67.66 & 69.97 & 67.11 & 67.24 \\
\cline{2-12}
& Avg Ensemble & \multicolumn{5}{c|}{70.80} & \multicolumn{5}{c|}{68.92} \\
\cline{2-12}
& Our RMB    & \multicolumn{5}{c|}{71.04} & \multicolumn{5}{c|}{68.70} \\
\hline
\multirow{3}{*}{HSIC}
& Individual RMs      & 73.40 & 70.73 & 69.29 & 68.05 & 73.84 & 70.08 & 69.95 & 67.58 & 66.97 & 71.62 \\
\cline{2-12}
& Avg Ensemble & \multicolumn{5}{c|}{72.04} & \multicolumn{5}{c|}{69.40} \\
\cline{2-12}
& Our RMB     & \multicolumn{5}{c|}{74.58} & \multicolumn{5}{c|}{71.65} \\
\hline
\end{tabular}%
}
\caption{Ensemble vs Boosting with different diversity strategies. We report the performance of individual RMs (index from 0 to 4).}
\label{tab:ablation_table}
\end{table*}

%% file: feature_correlation.tex
\begin{figure*}[t]
  \centering
  \captionsetup[subfigure]{skip=2pt,margin=0pt,labelformat=parens}
  \begin{subfigure}[t]{0.38\textwidth}
    \includegraphics[width=\linewidth]{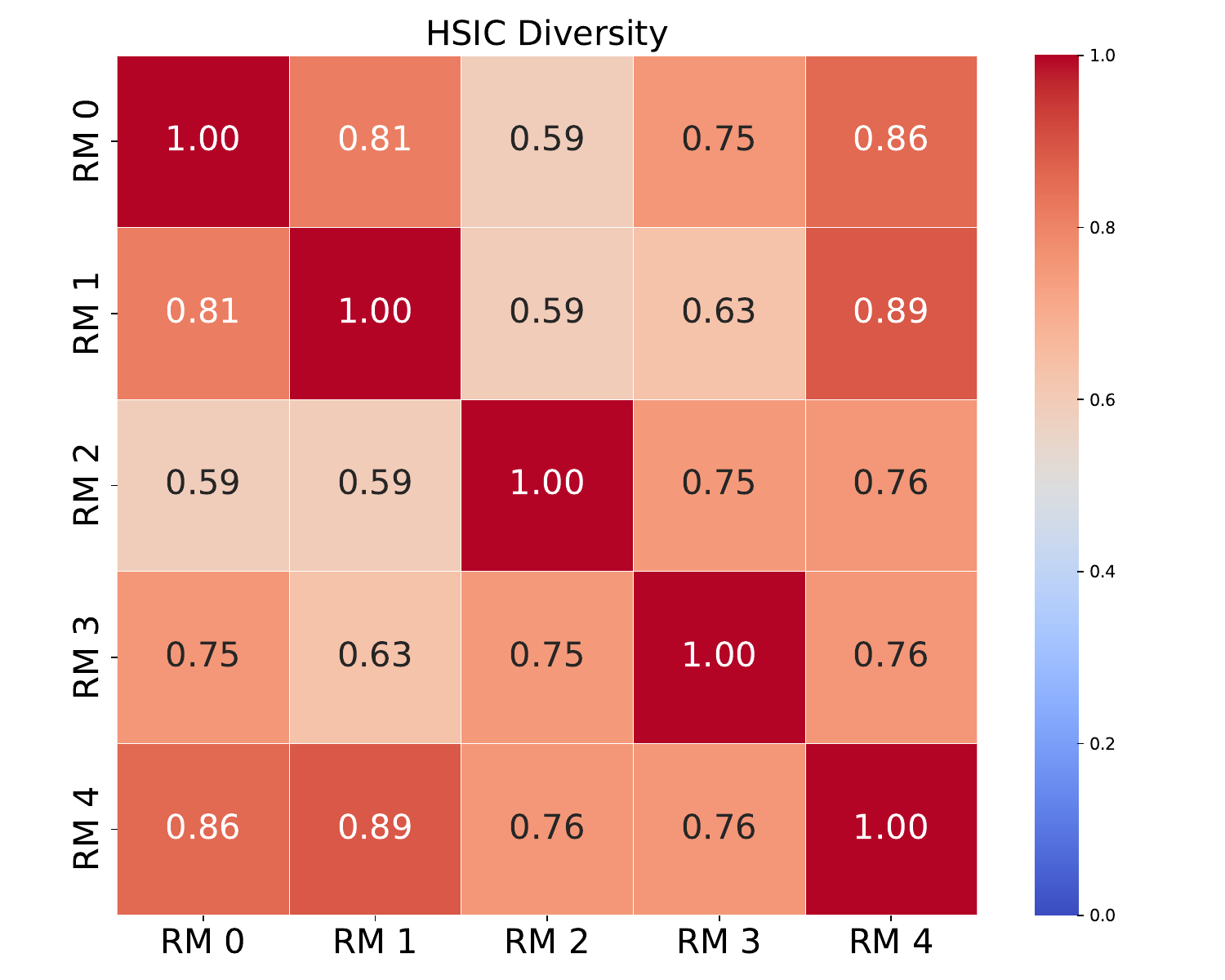} 
  \end{subfigure}
  \begin{subfigure}[t]{0.38\textwidth}
    \includegraphics[width=\linewidth]{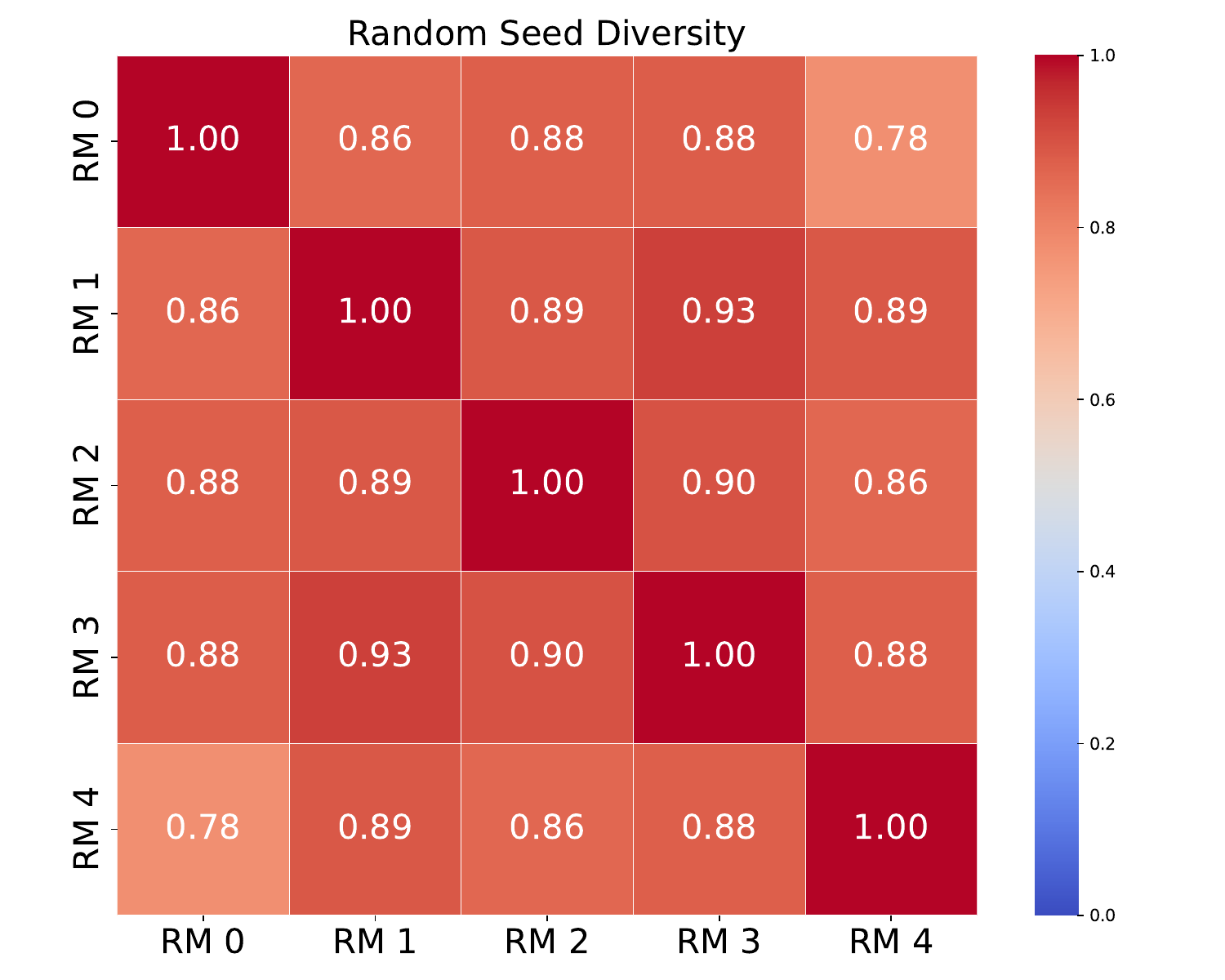}
  \end{subfigure}
  \caption{Correlation heat map among individual Reward Models}
  \label{fig:correlation}
\end{figure*}

%% file: feature_weights.tex
\begin{table}[t]
\centering
\renewcommand{\arraystretch}{1.0} 
 \resizebox{0.5\linewidth}{!}{%
\begin{tabular}{lc lc}
\toprule
\multicolumn{2}{c}{\textbf{HSIC Diversity}} & \multicolumn{2}{c}{\textbf{Random Seeds Diversity}} \\
\cmidrule(r){1-2} \cmidrule(l){3-4} 
\textbf{Model} & \textbf{Total Gain} & \textbf{Model} & \textbf{Total Gain} \\
\midrule
RM 0 & 29.75\% & RM 1 & 70.13\% \\
RM 4 & 27.39\% & RM 0 & 12.39\% \\
RM 1 & 26.78\% & RM 4 & 10.99\% \\
RM 2 & 8.55\%  & RM 2 & 5.82\%  \\
RM 3 & 7.53\%  & RM 3 & 0.66\%  \\
\bottomrule
\end{tabular}
}
\caption{Total Gain ratio of individual RMs.}
\label{tab:diversity_gain}
\end{table}

%% file: robustness.tex
{\captionsetup{skip=10pt, belowskip=-0pt}%
\begin{figure*}[t]
  \centering
  \captionsetup[subfigure]{skip=2pt,margin=0pt,labelformat=parens}

  \begin{subfigure}[t]{0.325\textwidth}
    \includegraphics[width=\linewidth]{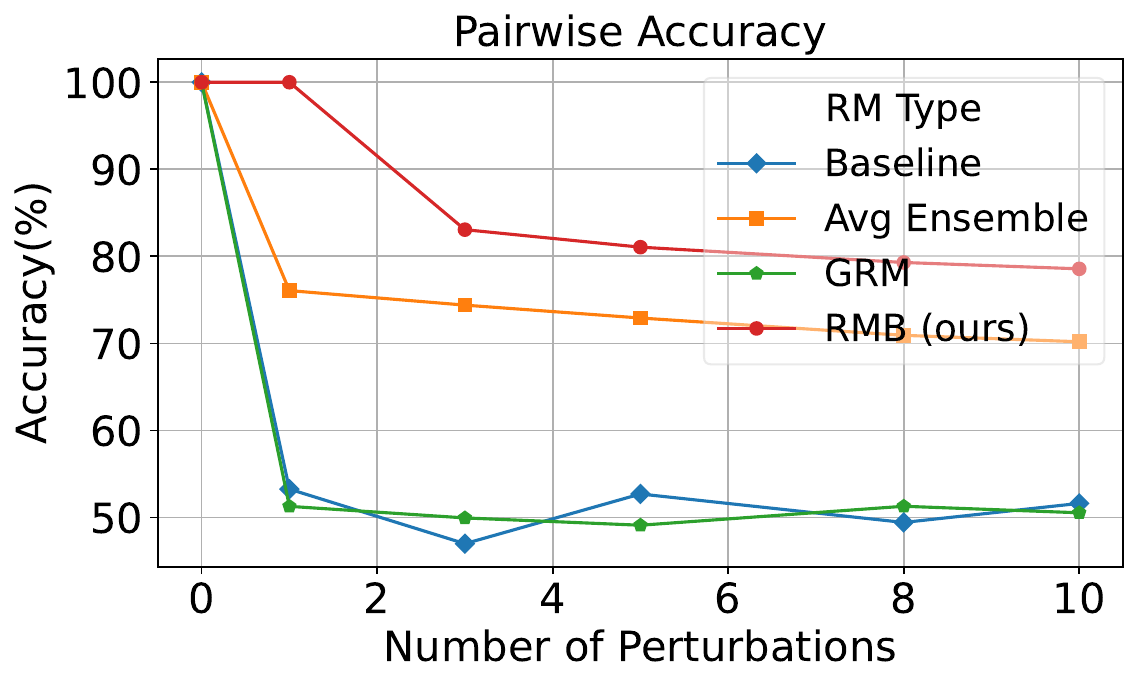} 
  \end{subfigure}
  \begin{subfigure}[t]{0.325\textwidth}
    \includegraphics[width=\linewidth]{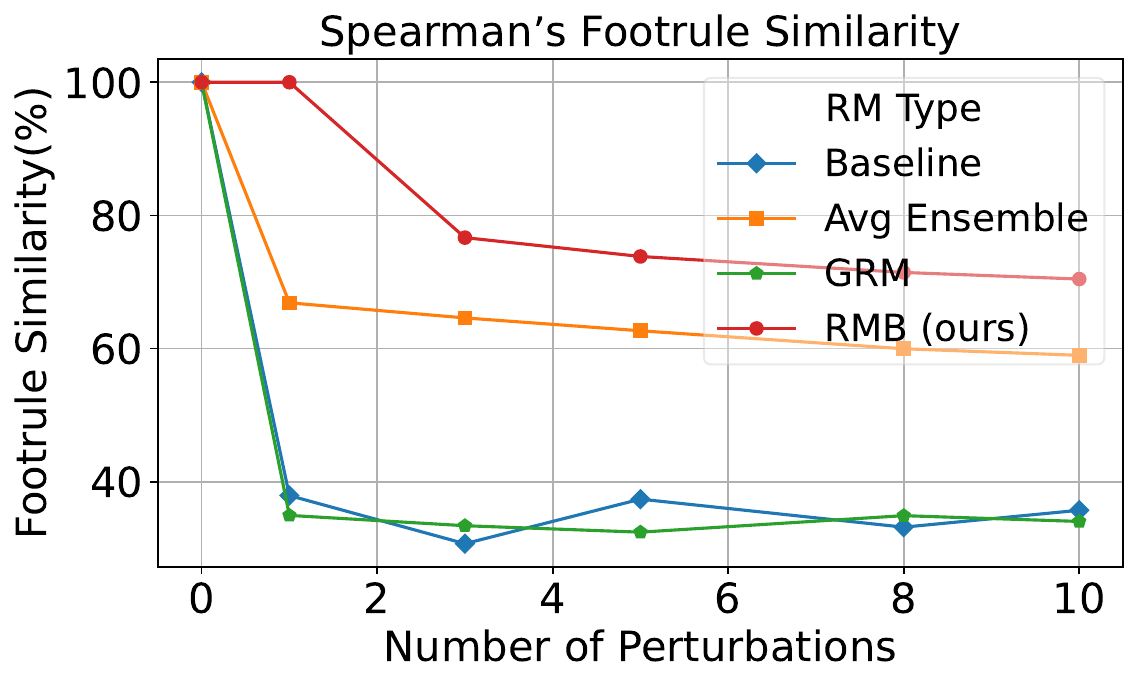}
  \end{subfigure}
  \begin{subfigure}[t]{0.325\textwidth}
    \includegraphics[width=\linewidth]{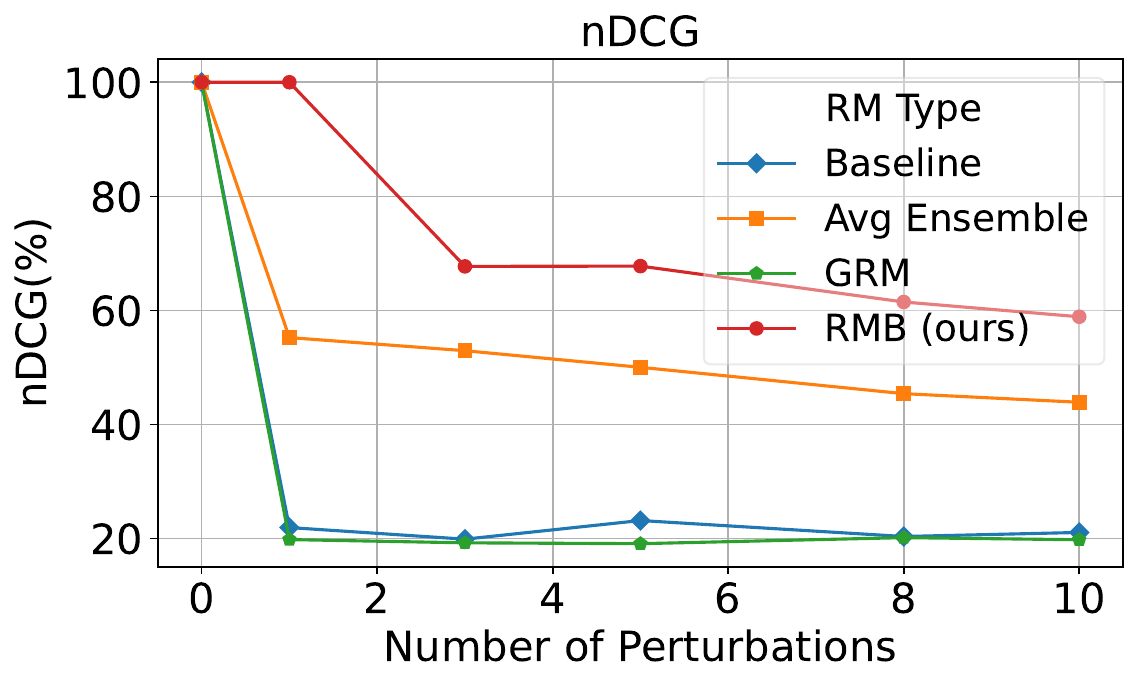}
  \end{subfigure}

  \caption{Rank order accuracy.}
  \label{fig:robustness_plt}
\end{figure*}
}

%% file: efficiency.tex
\begin{table}[t]
\centering
\resizebox{0.95\linewidth}{!}{
\small
\begin{tabular}{llcccc}
\toprule
\textbf{Latency Type} & \textbf{Component/Model} & \textbf{Mean} & \textbf{p50} & \textbf{p90} & \textbf{p99} \\
\midrule
\multirow{3}{*}{Breakdown (s/batch)}
& Base RMs 
& $19.7915$ & $19.7442$ & $20.2839$ & $20.3000$ \\
& Avg Ensemble aggregator
& $5.92\!\times\!10^{-5}$ & $5.67\!\times\!10^{-5}$ & $6.09\!\times\!10^{-5}$ & $1.398\!\times\!10^{-4}$ \\
& RMB aggregator
& $1.7458\!\times\!10^{-3}$ & $1.6997\!\times\!10^{-3}$ & $1.9010\!\times\!10^{-3}$ & $2.5694\!\times\!10^{-3}$ \\
\midrule
\multirow{2}{*}{End-to-End (s/batch)}
& Avg Ensemble
& $19.7915$ & $19.7443$ & $20.2840$ & $20.3001$ \\
& RMB
& $19.7932$ & $19.7460$ & $20.2857$ & $20.3016$ \\
\bottomrule
\end{tabular}}
\caption{Inference-efficiency summary.}
\label{tab:efficiency}
\end{table}

%% file: param_control.tex
\begin{table}[t]
\centering
\resizebox{0.75\linewidth}{!}{%
\begin{tabular}{lc ccccc} 
\toprule
\textbf{Reward Model} & \textbf{Total Params.} & \textbf{Average} & \textbf{Chat} & \textbf{Chat-Hard} & \textbf{Safety} & \textbf{Reasoning} \\
\midrule
RMB (N=3) & 6B & \textbf{87.8} & 98.9 & 75.6 & 87.3 & \textbf{89.1} \\
GPT-4o-2024-08-06 & - & 86.7 & 96.1 & \textbf{76.1} & 88.1 & 86.6 \\
GPT-4-0125-preview & - & 86.0 & 95.3 & 74.3 & 87.6 & 86.9 \\
FsfairX-LLaMA3-RM & 8B & 84.4 & \textbf{99.4} & 65.1 & 86.8 & 86.4 \\
Starling-RM & 34B & 82.7 & 96.9 & 57.2 & \textbf{88.2} & 88.5 \\
\bottomrule
\end{tabular}
}
\caption{Results of different reward models in a budget-constrained scenario.}
\label{tab:reward_model_comparison}
\end{table}

%% file: implement_details.tex
\begin{table}[bt]%
	\centering%

	\centering
	\resizebox{0.55\textwidth}{!}{
		\begin{tabular}{ll}%
			\toprule
			\multicolumn{2}{c}{\textbf{Basic Setups}}                   \\                                                        
			\midrule                                                                                
			Base models         & \href{https://huggingface.co/google/gemma-2b-it}{Gemma-2b-it}                                       \\     
            Quantization for training & bf16 \\
            Fine-tuning strategy & LoRA                                                                                                                                                                                     \\
            LoRA $r$             & 32\\
 			LoRA alpha           & 64                                                                                                                                                                                     \\
			LoRA dropout         & 0.05                                                                                                                                                                                                                                                            \\
			Optimizer            & Adamw\_hf           \\   
   			Batch size           & 16      \\
            Learning rate     & $1\times 10^{-5}$ \\
            Learning rate scheduler & cosine \\
            Warmup ratio & 0.03 \\

			\midrule
			\multicolumn{2}{c}{\textbf{RMB Setups}}              \\       
			\midrule
             HSIC ratio $\lambda$ & 0.1 \\
             Max depth & 5 \\
             Max round of boosting & 256 \\
             L2 regularizer & 0.1 \\

            \midrule
			\multicolumn{2}{c}{\textbf{PPO Setups} \citep{schulman2017proximal} }                                                        \\
			\midrule
			KL regularization               & 0.0                          \\
			Epochs               & 1                       \\          
            Learning rate        & $1\times 10^{-5}$  \\
            Lambda for GAE    & 0.95 \\
            Gamma             & 1 \\
            Clip range         & 0.2 \\
            Number of optimization epochs per batch & 4 \\
            Number of tokens during generation & 512 \\

			\bottomrule
		\end{tabular}
	}
	\caption{Key implementation details of our main experiments.}
	\label{tab:exp_details}
\end{table}%

%% file: 400k_main.tex
\definecolor{avgcol}{gray}{0.92}
\newcolumntype{A}{>{\columncolor{avgcol}}c} 

\begin{table*}[t]
\centering
\scriptsize
\setlength{\tabcolsep}{5pt}
\renewcommand{\arraystretch}{1.15}
\begin{tabular}{lcccAcccc}
\toprule
\multirow{2}{*}{\textbf{Reward model}} &
\multirow{2}{*}{\textbf{Unified-Feedback}} &
\multirow{2}{*}{\textbf{HHH-Align}} &
\multirow{2}{*}{\textbf{MT-Bench}} &
\multicolumn{5}{c}{\textbf{RewardBench}} \\
\cmidrule(lr){5-9}
& & & & \multicolumn{1}{c}{Avg.} & Chat & Chat-Hard & Safety & Reasoning \\
\midrule
Frozen RM        & 63.8  & 66.4  & 69.5  & x     & x     & x     & x     & x     \\
Baseline RM      & 72.1  & 73.4  & 71.2  & 68.2  & 95.5  & 38.0  & 73.8  & 65.3  \\
Label Smooth   & 71.5  & 72.1  & 71.2  & 70.6  & 94.4  & 37.3  & 73.2  & \textbf{77.4}  \\
Margin Loss    & 72.0  & 75.0  & 72.6  & 70.2  & 95.8  & 38.4  & 73.9  & 72.5  \\
GRM           & 73.2  & 79.8  & 73.4  & 70.8  & 97.8  & 42.1  & 77.9  & 65.2  \\
Avg Ensemble ($N=3$)      & 72.8  & 76.8  & 73.7  & 71.0  & \textbf{98.0}  & 37.5  & 77.3  & 71.3  \\
Avg Ensemble ($N=5$)      & 73.2  & 79.5  & 73.8  & 71.6  & 95.3  & 43.4  & 78.3  & 71.4  \\
\midrule
RMB ($N=3$)           & 74.4 & 81.8 & 75.2 & 72.0 & 95.6 & \textbf{44.1} & 78.1 & 71.8 \\
RMB ($N=5$)           & 75.1 & 82.6 & 75.6 & 72.2 & 91.9 & 39.4 & \textbf{80.7} & 73.7 \\
RMB ($N=8$)           & \textbf{75.9} & \textbf{83.0} & \textbf{76.6} & \textbf{73.3} & 95.1 & 43.5 & 76.6 & 74.5 \\
RMB ($N=10$)          & 74.9 & 82.1 & 76.2 & 72.6 & 96.9 & 42.8 & 77.9 & 73.3 \\
\bottomrule
\end{tabular}
\caption{Preference prediction performance on all Reward Models learned using 400K data mode of Unified-Feedback.  Results of competing approaches are directly sourced from \citet{yang2024regularizing}.  $N$ means the number of reward models used for ensembling or boosting.}
\label{tab:reward_performance_400k}
\end{table*}

%% file: win_rate_gpt4o.tex
\begin{table*}[t]
\vspace*{-0pt}
\centering
\resizebox{0.95\linewidth}{!}{
\begin{tabular}{c c c c c c c c c c}
\toprule
\multirow{2}{*}{\textbf{Approach}} &
\multirow{2}{*}{\textbf{Opponent}} &
\multicolumn{4}{c}{\textbf{Pairwise Preference Comparison} (in \%)} &
\multicolumn{4}{c}{\textbf{Pointwise Preference Comparison} (in \%)}\\
\cmidrule(lr){3-6}\cmidrule(lr){7-10}
& & \textbf{Win$\uparrow$} & \textbf{Tie} & \textbf{Lose$\downarrow$} & \boldmath{$\Delta$} &
\textbf{Win$\uparrow$} & \textbf{Tie} & \textbf{Lose$\downarrow$} & \boldmath{$\Delta$} \\
\midrule
\multirow{3}{*}{\textbf{RMB} ($N=3$)} 
 & Baseline RM   & 59.54 & 5.42 & 35.04 & $\uparrow$ \textbf{24.50} & 47.28 & 20.19 & 32.53 & $\uparrow$ \textbf{14.75} \\
 & Avg Ensemble ($N=3$)  & 53.15 & 7.93 & 38.92 & $\uparrow$ \textbf{14.23} & 39.40 & 30.91 & 29.69 & $\uparrow$ \textbf{9.71} \\
 & GRM           & 57.38 & 9.37 & 33.25 & $\uparrow$ \textbf{24.13} & 41.46 & 27.44 & 31.10 & $\uparrow$ \textbf{10.36}  \\
\bottomrule
\end{tabular}
}
\caption{LLM-as-judge (GPT-4o) of policies learned via RLHF.}
\label{tab:llm_eval_4o}
\end{table*}

%% file: tradeoff_plot.tex
\begin{figure*}[t]
  \centering
  \captionsetup[subfigure]{skip=2pt,margin=0pt,labelformat=parens}
  \begin{subfigure}[t]{0.6\textwidth}
    \includegraphics[width=\linewidth]{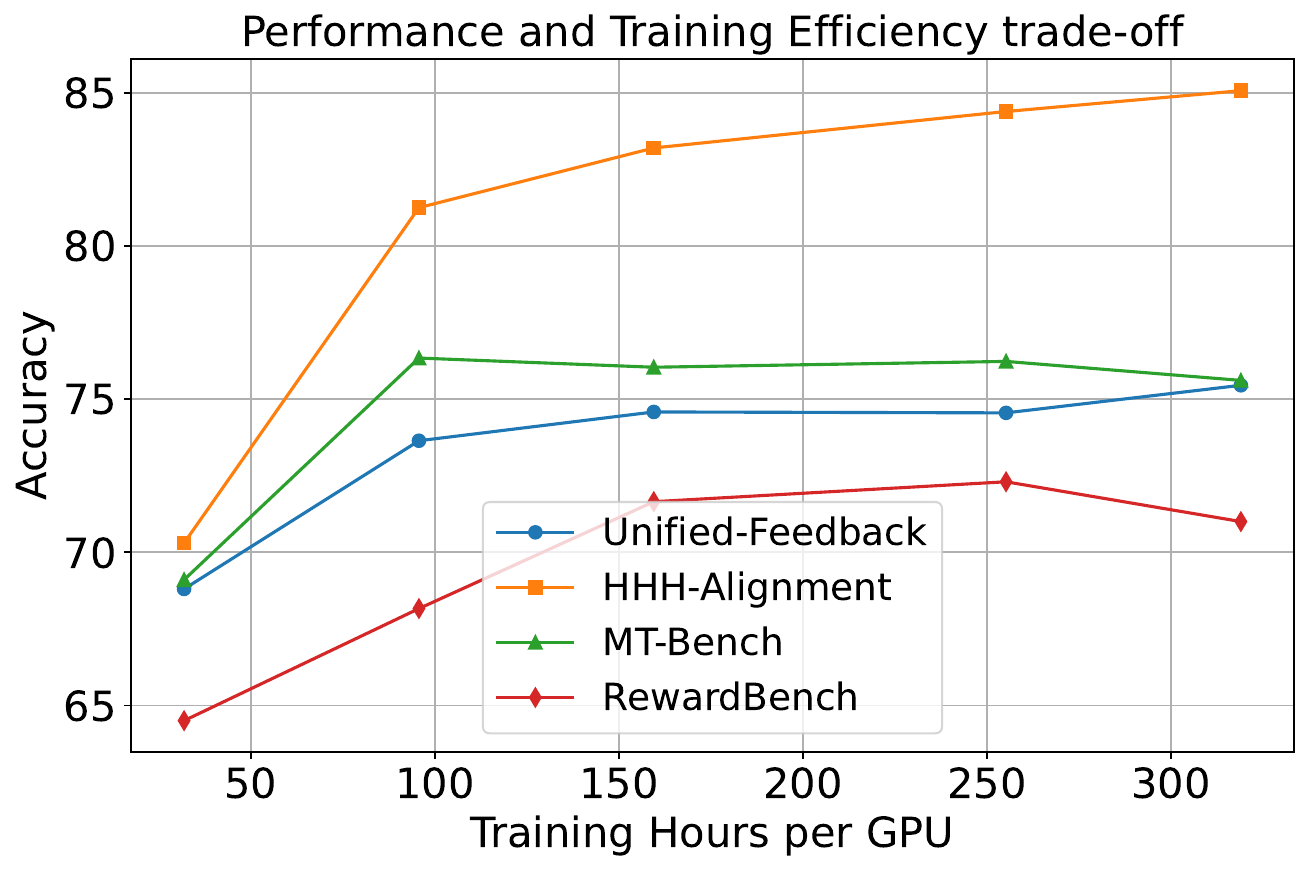} 
  \end{subfigure}
  \caption{Performance--training efficiency trade-off.}
  \label{fig:tradeoff}
\end{figure*}

%% file: sensitivity.tex
\begin{table}[h]
\centering
\begin{tabular}{c c c c}
\hline
$\lambda_{\text{HSIC}}$ & Avg. Indiv. RM Perf. & Avg. Pairwise Corr. & RMB Perf. \\
\hline
0.001 & $70.7 \pm 0.8$ & 92\% & 71.7 \\
0.01  & $71.3 \pm 1.1$ & 86\% & 72.8 \\
0.1   & $71.1 \pm 2.5$ & 73\% & \textbf{74.6} \\
1     & $68.8 \pm 3.0$ & 59\% & 71.2 \\
10    & $61.7 \pm 3.7$ & 51\% & 66.5 \\
\hline
\end{tabular}
\caption{Sensitivity analysis of the $\lambda_{\text{HSIC}}$. We report the average performance of individual Reward Models, the average pairwise correlation between them, and the final RMB performance.}\label{tab:sensitivity}
\end{table}